\documentclass[letterpaper]{article} 
\usepackage{aaai2026}  
\usepackage{times}  
\usepackage{helvet}  
\usepackage{courier}  
\usepackage[hyphens]{url}  
\usepackage{graphicx} 
\usepackage{natbib}  
\usepackage{caption} 
\usepackage{algorithm}
\usepackage{algorithmic}
\usepackage{color}

\usepackage{multirow}
\usepackage{pifont}
\usepackage{booktabs}
\usepackage{makecell}
\usepackage[table,xcdraw,svgnames]{xcolor}
\usepackage{tcolorbox}
\usepackage{colortbl}
\usepackage{arydshln}
\usepackage{subcaption}
\newcommand{\cmark}{\textcolor{green}{\ding{51}}}
\newcommand{\xmark}{\textcolor{red}{\ding{55}}}

\usepackage{newfloat}
\usepackage{listings}
\DeclareCaptionStyle{ruled}{labelfont=normalfont,labelsep=colon,strut=off} 
\floatstyle{ruled}
\newfloat{listing}{tb}{lst}{}
\floatname{listing}{Listing}
\title{CrossVid: A Comprehensive Benchmark for Evaluating Cross-Video Reasoning in Multimodal Large Language Models}
\author{
    Jingyao Li\textsuperscript{\rm 1}\equalcontrib, 
    Jingyun Wang\textsuperscript{\rm 1}\equalcontrib, 
    Molin Tan\textsuperscript{\rm 1}\equalcontrib, 
    Haochen Wang\textsuperscript{\rm 1}, 
    Cilin Yan\textsuperscript{\rm 1}, \\
    Likun Shi\textsuperscript{\rm 1}, 
    Jiayin Cai\textsuperscript{\rm 1}\thanks{Corresponding author.}, 
    Xiaolong Jiang\textsuperscript{\rm 1}, 
    Yao Hu\textsuperscript{\rm 1}
}
\affiliations{
    \textsuperscript{\rm 1}Xiaohongshu Inc., China\\
    \{lijingyao, tanmolin, shilikun, laige\}@xiaohongshu.com, \{clyanhh, yaoohu\}@gmail.com, \\1411249598@qq.com, h.wang3@uva.nl, caijy18@tsinghua.org.cn
}

\usepackage{bibentry}

\begin{document}

\maketitle

\begin{abstract}
Cross-Video Reasoning (CVR) presents a significant challenge in video understanding, which requires simultaneous understanding of multiple videos to aggregate and compare information across groups of videos.
Most existing video understanding benchmarks focus on single-video analysis, failing to assess the ability of multimodal large language models (MLLMs) to simultaneously reason over various videos.
Recent benchmarks evaluate MLLMs' capabilities on multi-view videos that capture different perspectives of the same scene.
However, their limited tasks hinder a thorough assessment of MLLMs in diverse real-world CVR scenarios.
To this end, we introduce \textbf{CrossVid}, the first benchmark designed to comprehensively evaluate MLLMs' spatial-temporal reasoning ability in cross-video contexts.
Firstly, CrossVid encompasses a wide spectrum of hierarchical tasks, comprising four high-level dimensions and ten specific tasks, thereby closely reflecting the complex and varied nature of real-world video understanding.
Secondly, CrossVid provides 5,331 videos, along with 9,015 challenging question-answering pairs, spanning single-choice, multiple-choice, and open-ended question formats.
Through extensive experiments on various open-source and closed-source MLLMs, we observe that Gemini-2.5-Pro performs best on CrossVid, achieving an average accuracy of 50.4\%. 
Notably, our in-depth case study demonstrates that most current MLLMs struggle with CVR tasks, primarily due to their inability to integrate or compare evidence distributed across multiple videos for reasoning.
These insights highlight the potential of CrossVid to guide future advancements in enhancing MLLMs’ CVR capabilities. 
\end{abstract}

\begin{links}
\link{Datasets}{https://github.com/chuntianli666/CrossVid}
\end{links}

\section{Introduction}

\begin{table*}[!ht]
\centering
\setlength{\tabcolsep}{2mm}
{\small

\begin{tabular}{lccccccccc}
\toprule[1pt]
\textbf{Benchmarks}       & \textbf{\#Videos} & \textbf{\#QA pairs} & \textbf{Len. (s)} & \textbf{\#Tasks}  &\textbf{Anno.}& \makecell{\textbf{Closed-} \\ \textbf{ended}} & \makecell{\textbf{Open-} \\ \textbf{ended}} &  \makecell{\textbf{Multi-} \\ \textbf{video}}  & \makecell{\textbf{Multi-} \\ \textbf{view}} \\
\midrule
TVQA \cite{tvqa}       & 2,179     & 15,253      & 11  & 3 & M     &  \cmark  & \xmark   & \xmark                    & \xmark          \\
MVBench \cite{mvbench}         & 3,641& 4,000& 16                  &          20&A& \cmark           & \xmark          & \xmark                    & \xmark          \\
ActivityNet-QA \cite{activitynet}   &          5,800&            58,000&                     180&          4&M&             \xmark                    &            \cmark          & \xmark                    & \xmark          \\
NExT-QA \cite{nextqa}          & 5,440& 52,044& 44&          2&M& \cmark           & \cmark          & \xmark                    & \xmark          \\
LongVideoBench \cite{longvideobench}   & 3,763& 6,678& 473                 &          17&M& \cmark           & \xmark          & \xmark                    & \xmark          \\
MMVU \cite{mmvu}             & 1,529& 3,000& 51                & 27 &M& \cmark           & \cmark          & \xmark                    & \xmark          \\
Video-MME \cite{videomme}        & 900      & 2,700& 1,017                & 12       &M& \cmark           & \xmark          & \xmark                    & \xmark          \\
MLVU \cite{mlvu}             & 1,730& 3,102& 930                 & 9        &M+A& \cmark           & \cmark          & \xmark                    & \xmark          \\
Ego-Exo4D \cite{ego-exo4d}                & 5,035 & - & 156 & 4 & M & \xmark & \xmark &\cmark &\cmark \\
EgoExoLearn \cite{egoexolearn}& 747& -& -& 4& M& \cmark           & \cmark           & \cmark           &\cmark           \\
All-Angles Bench \cite{all-angels-bench} &          90 scenes& 2,132       & -                   & 6        &M& \cmark           & \xmark          & \cmark                    & \cmark          \\
\rowcolor{gray!10}
\textbf{CrossVid~(Ours)}             & 5,331     & 9,015      &          215   & 10       & M+A & \cmark           & \cmark          & \cmark                    & \cmark          \\ 
\arrayrulecolor{black}
\bottomrule[1pt]
\end{tabular}
}
\small\caption{Comparison of CrossVid with existing benchmarks, including \textbf{\#Videos} (number of videos), \textbf{\#QA pairs} (number of QA pairs), \textbf{Len.} (average length of videos in seconds), \textbf{\#Tasks} (number of distinct tasks), \textbf{Anno.} (annotation pipeline, M: manual, A: automated), presence of \textbf{Open-ended} and \textbf{Closed-ended} tasks, presence of \textbf{Multi-video reasoning} and \textbf{Multi-view reasoning}. }
\label{table_comparison}
\end{table*}

\begin{figure}[t]
\centering
\includegraphics[width=0.9\columnwidth]{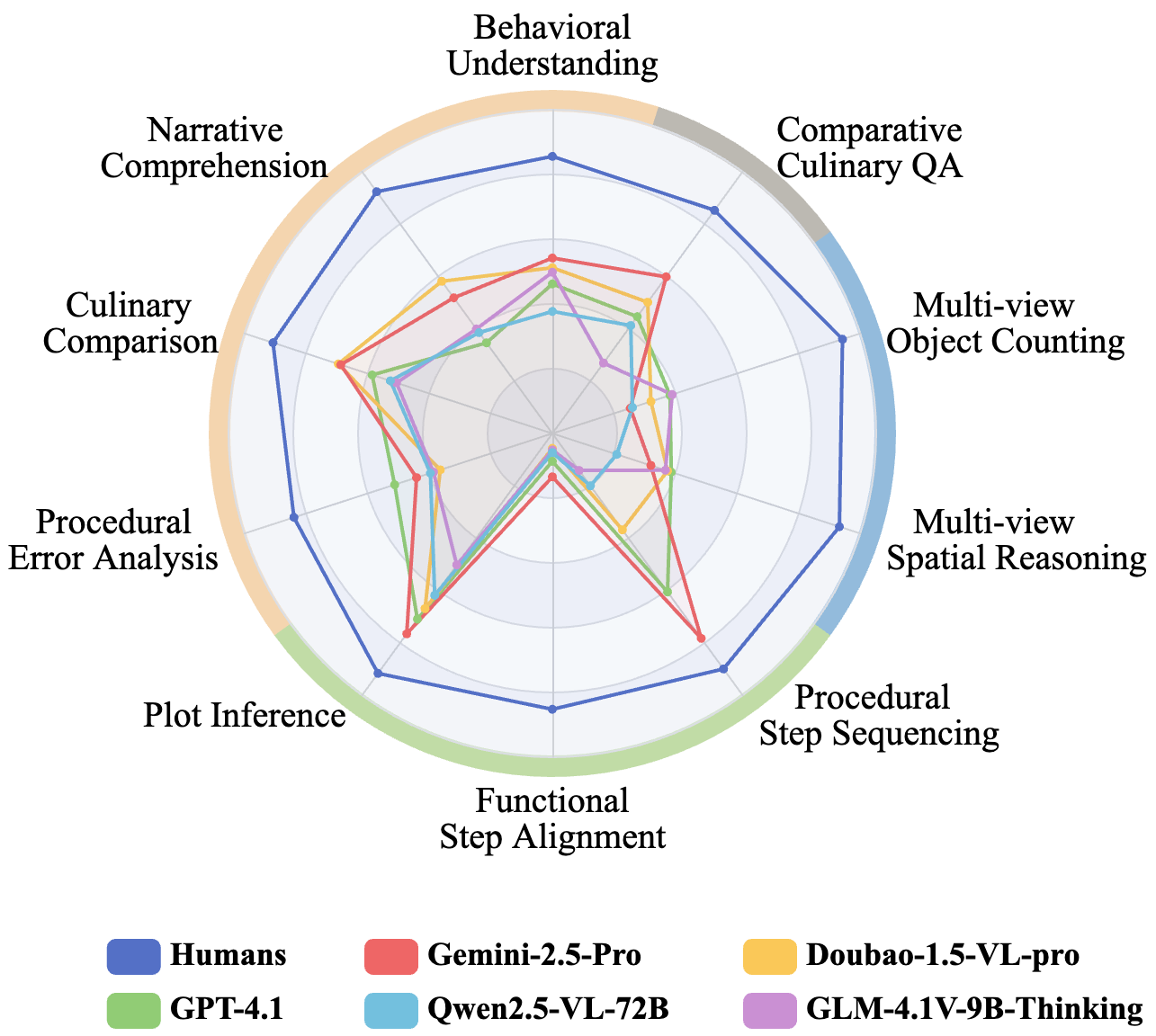}
\caption{Performance of MLLMs on CrossVid.}
\label{fig_radar}
\end{figure}

With the rapid development of multimodal large language models (MLLMs) \cite{qwen2.5, gpt4o, gemini-25-pro}, 
video reasoning \cite{video-llava, sharegpt4video, video-xl, videollama} emerges as an important testbed for evaluating the reasoning capabilities of MLLMs.
However, most existing benchmarks \cite{videomme, nextqa, activitynet} for video reasoning primarily focus on single-video analysis, severely restricting the assessment of MLLMs' reasoning capabilities on more complex tasks that span multiple videos.

Cross-Video Reasoning (CVR) is a challenging yet essential task within the domain of video reasoning.
Aiming to aggregate and compare information across videos, CVR requires to simultaneously understand multiple videos.
A recent research, All-Angles Bench~\cite{all-angels-bench}, evaluates MLLMs' performance on groups of multi-view videos, where each video captures a different perspective of the same scene.
However, the task of All-Angles Bench is limited to multi-view videos showing the same scene, hindering a thorough assessment of MLLMs’ CVR abilities across the diverse and complex scenarios in the real world.

To this end, we propose \textbf{CrossVid}, the first video reasoning benchmark designed to advance from previous single-query, single-video paradigms to single-query, multi-video understanding, and to comprehensively evaluate MLLMs' spatial-temporal reasoning ability for CVR.
CrossVid features a wide range of hierarchical tasks, reflecting the diversity of real-world video understanding scenarios.
It consists of 4 high-level dimensions, including comparative analysis, temporal understanding, multi-view reasoning, and free-form QA, encompassing a total of 10 specific tasks.
CrossVid consists of 5,331 videos and 9,015 challenging QA pairs, covering single-choice, multiple-choice, and open-ended question formats.
On average, each query requires MLLMs to comprehend approximately 770 seconds of video content.
To ensure precise annotation, we develop a semi-automated annotation pipeline and employ 10 expert annotators to facilitate the construction.

Extensive experiments on CrossVid are conducted on various
representative closed-source~\cite{gpt4o, gemini-25-pro} and open-source MLLMs~\cite{qwen2.5, internvl3}, ranging from 7B to 78B parameters and diverse architectures.
As shown in Figure~\ref{fig_radar}, while current MLLMs excel in single-video tasks, they still struggle with CVR.
Notably, Gemini-2.5-Pro achieves the best average accuracy of 50.4\%.
Furthermore, we provide multiple key insights based on the experimental results, which further verify that our proposed CrossVid establishes new pathways for MLLMs’ future advancements in video reasoning.
Our detailed case studies and ablation experiments confirm that, despite ongoing advances, MLLMs still struggle to aggregate and compare evidence distributed across videos—a fundamental capability for real-world CVR.

In summary, our main contributions are:

\begin{itemize} 
\item We propose \textbf{CrossVid}, the first benchmark to systematically evaluate MLLMs’ CVR capability. CrossVid incorporates hierarchical tasks spanning four high-level dimensions and ten specific tasks. The dataset is constructed with a semi-automated annotation pipeline under rigorous quality controls. It contains 9,015 high-quality QA pairs and 5,331 videos, including both closed-ended and open-ended question formats.
\item We conduct extensive experiments on 22 representative closed-source and open-source MLLMs. Our detailed case analysis and ablation studies offer critical insights into current limitations of MLLMs for CVR, paving the way for improvements in future development in video understanding for MLLMs. \end{itemize}

\section{Related Work}
\subsection{Video Understanding MLLMs}
MLLMs have demonstrated remarkable advancements in video understanding by integrating visual encoders with large language models and fine-tuning on downstream tasks~\cite{videollama,video-llava,llava-video}. Previous works mainly focus on single-video understanding, utilizing key frame selection~\cite{key-frame1, key-frame2} and token compression~\cite{video-xl, moviechat} to accomplish tasks such as video captioning, action recognition, and long-video comprehension. Notably, models like Qwen2.5-VL~\cite{qwen2.5} have shown the ability to process hour-long videos with improved temporal reasoning.

However, despite these advances, most existing open-source MLLMs remain untrained for comprehensive CVR (\emph{i.e.}, joint inference across multiple input videos). A few recent works~\cite{reilly2025my} begin to explore cross-perspective understanding, but do not generalize to broader multi-video settings.

\begin{figure*}[t]
\centering
\includegraphics[width=\textwidth]{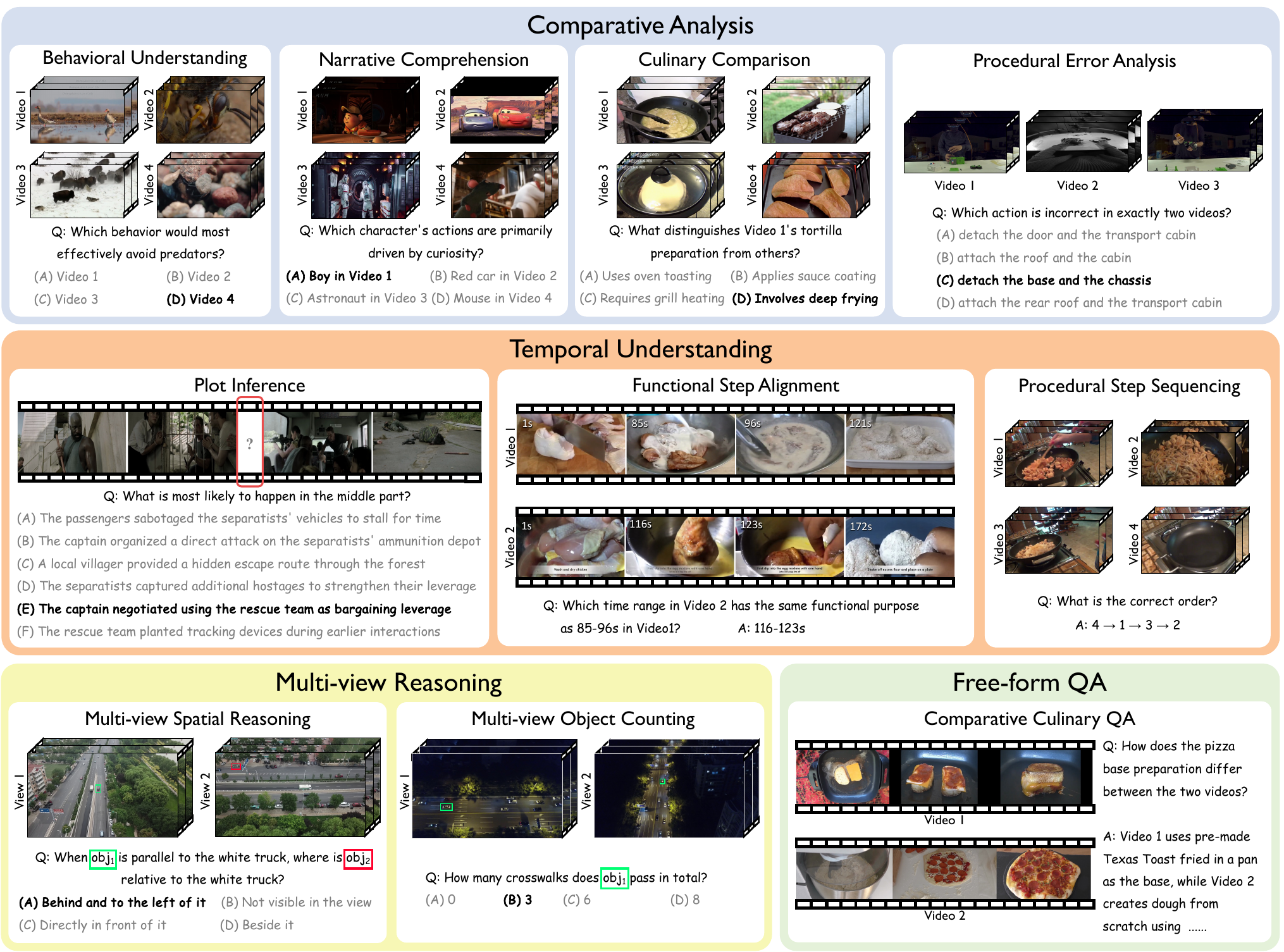}
\small\caption{Overview of CrossVid. It evaluates MLLMs' CVR capability on 4 dimensions: comparative analysis, temporal understanding, multi-view reasoning, and free-form QA. It contains 10 distinct tasks: behavioral understanding (\textbf{BU}), narrative comprehension (\textbf{NC}), culinary comparison (\textbf{CC}), procedural error analysis (\textbf{PEA}), plot inference (\textbf{PI}), functional step alignment (\textbf{FSA}), procedural step sequencing (\textbf{PSS}), multi-view spatial reasoning (\textbf{MSR}), multi-view object counting (\textbf{MOC}) and comparative culinary QA (\textbf{CCQA}).}
\label{fig_examples}
\end{figure*}

\subsection{Video QA Benchmarks}
Video Question Answering (VQA) benchmarks have mostly focused on assessing models' abilities to understand and reason over single videos. Early works such as TVQA~\cite{tvqa} and ActivityNet-QA~\cite{activitynet} require understanding short or long video clips via closed- or open-ended questions. To target spatial and temporal reasoning, datasets like NExT-QA~\cite{nextqa} and LongVideoBench~\cite{longvideobench} are introduced. Some recent efforts, such as MVBench~\cite{mvbench} and Video-MME~\cite{videomme}, extend the diversity of covered tasks.

More recent works start to target multi-view reasoning. For example, EgoExoLearn~\cite{egoexolearn} evaluates across exocentric and egocentric views, and All-Angles Bench~\cite{all-angels-bench} introduces multi-view video QAs. However, their scale and task coverage are limited, and they primarily focus on specific domains or perspectives. 

To our knowledge, there is currently no large-scale, systematically annotated benchmark for general CVR. CrossVid is the first to comprehensively benchmark MLLMs on a diverse suite of CVR tasks, providing an important resource to spur future advances in this field.

\section{CrossVid Benchmark}
In this section, we first provide an overview of our CrossVid and its data curation process.
We then present our semi-automated annotation pipeline used for its construction.

\begin{figure*}[h]
\begin{subfigure}[b]{0.37\linewidth}
    \centerline{\includegraphics[height=6cm]{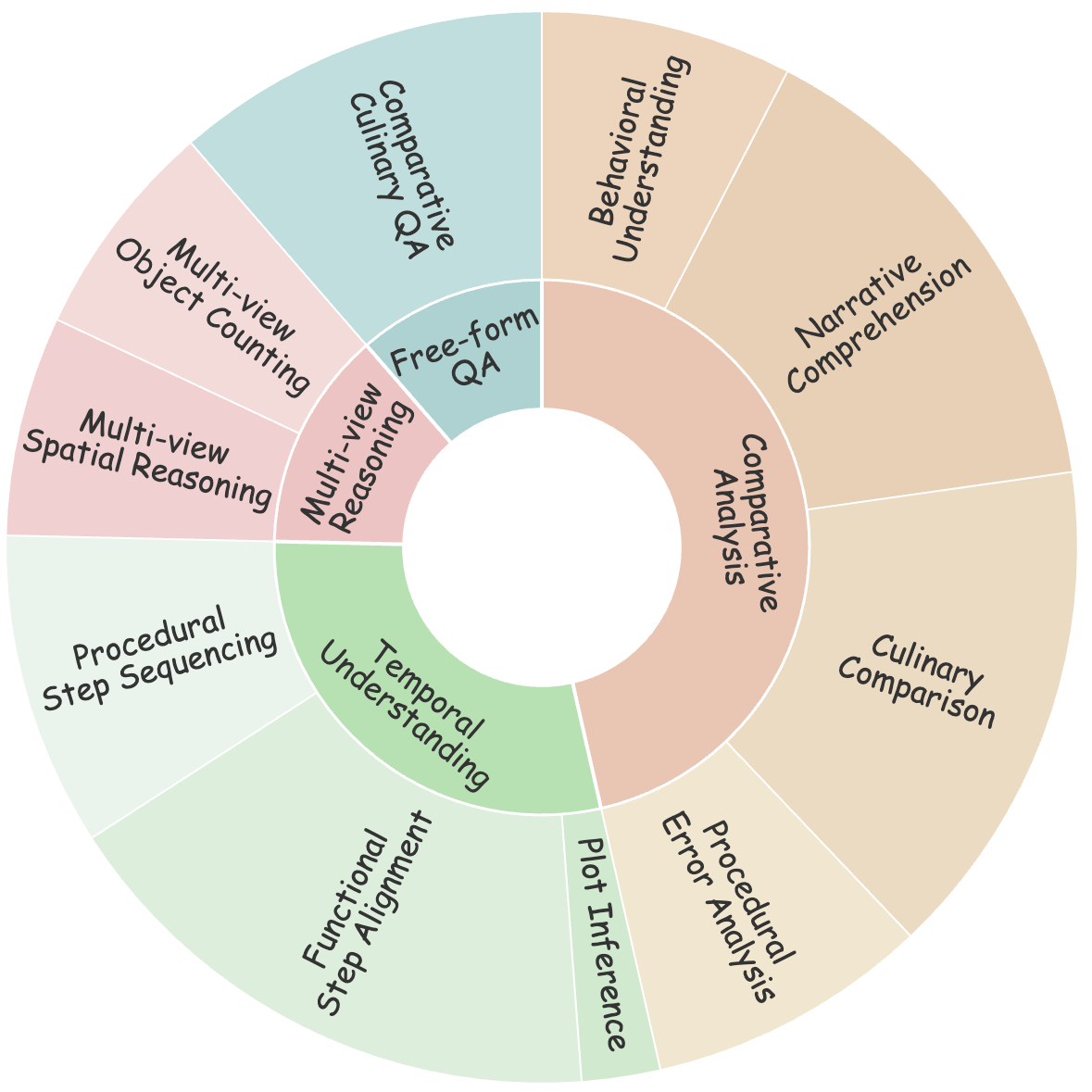}}
    \caption{Types of dimensions and tasks.}
    \label{fig:tasks}
\end{subfigure}
\begin{subfigure}[b]{0.37\linewidth}
    \centerline{\includegraphics[height=6cm]{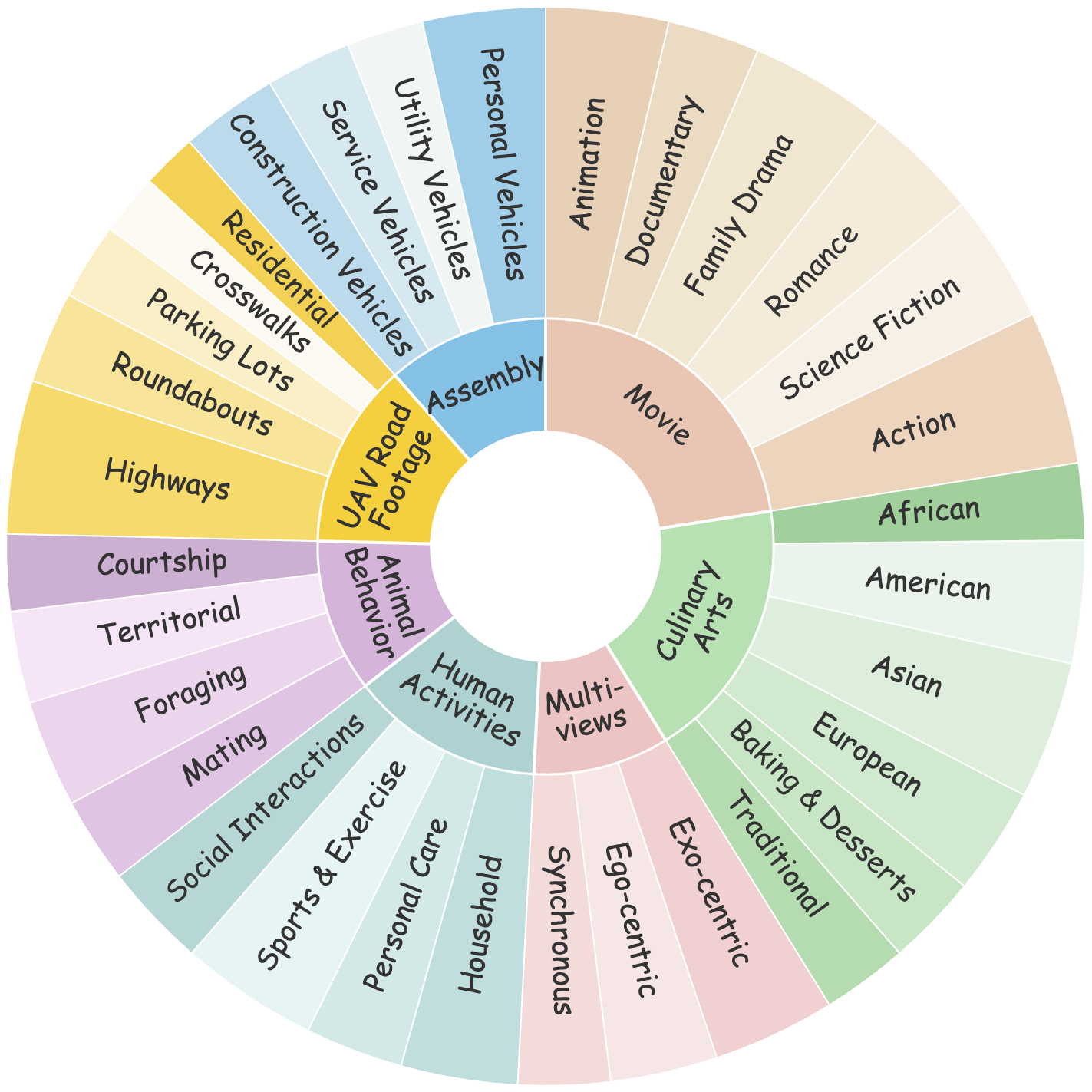}}
    \caption{Categories and genres of videos.}
    \label{fig:genres}
\end{subfigure}
\begin{subfigure}[b]{0.25\linewidth}
    \centerline{\includegraphics[height=6cm]{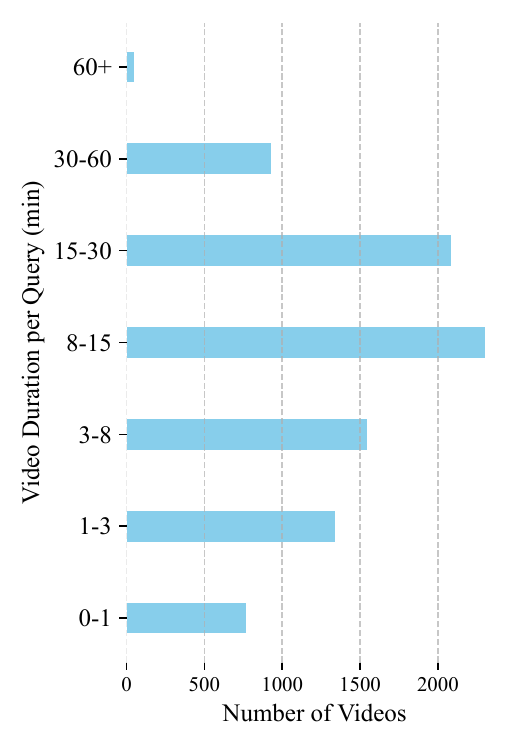}}
    \caption{Distribution of total video duration involved in each query.}
    \label{fig:videoduration}
\end{subfigure}
\caption{Statistical analysis of our CrossVid dataset. It consists of 4 high-level dimensions and 10 specific tasks, covering a wide range of video durations and video sources of 7 primary categories and 32 genres.}
\label{fig:benchmark_statistic}
\end{figure*}

\subsection{Overview}
\label{method_overview}
CrossVid is the first large-scale benchmark to systematically evaluate MLLMs' capabilities for CVR. 
This benchmark specifically assesses models' ability to integrate, compare, and reason over information from a group of related videos. 

\noindent \textbf{Video curation}
CrossVid consists of 5,331 video clips curated from six diverse, publicly available datasets: Animal Kingdom \cite{animalkingdom}, MovieChat-1K \cite{moviechat}, YouCook2 \cite{youcook}, VisDrone \cite{visdrone}, Charades \cite{charades}, and Assembly101 \cite{assembly101}.
With the various sources, CrossVid covers a wide range of video lengths and various degrees of visual complexity.
During video selection, we emphasize scenario diversity, action complexity, and inter-video correlation, ensuring that the resulting multi-video groups are both challenging and appropriate for CVR tasks.

\noindent \textbf{Hierarchical tasks}
Based on the curated video clips, we further propose 9,015 high-quality QA pairs.
Each piece in CrossVid is composed of a group of semantically related videos, a task-specific query targeting CVR, and a carefully verified reference answer.
As shown in Figure~\ref{fig_examples}, all queries in CrossVid form hierarchical tasks with 4 high-level dimensions, including comparative analysis, temporal understanding, multi-view reasoning, and free-form QA.
These 4 high-level dimensions are further divided into 10 distinct tasks, which are shown in Figure~\ref{fig:tasks}.

Besides, CrossVid encompasses 32 various genres in Figure~\ref{fig:genres}, fully capturing representative CVR scenarios encountered in the real world. Furthermore, CrossVid features a hierarchical distribution of video duration per query, ranging from 1 minute to over 1 hour, which is presented in Figure~\ref{fig:videoduration}.
More statistical analysis and details of the tasks in CrossVid can be found in the Appendix.

\noindent \textbf{Comparison with previous benchmarks}
Table \ref{table_comparison} summarizes the characteristics of existing VQA benchmarks. Compared with previous single-video understanding benchmarks, CrossVid innovatively introduces multi-video input and cross-video understanding. Compared with current multi-view benchmarks, CrossVid significantly extends the types of tasks, the formats of questions, and the application scenarios. Therefore, CrossVid is the first benchmark comprehensively evaluating MLLMs' CVR capability with a wide coverage of tasks and question formats.

\subsection{Data Annotation}
\label{method_annotation}

\begin{figure*}[!h]
\centering
\includegraphics[width=0.98\textwidth]{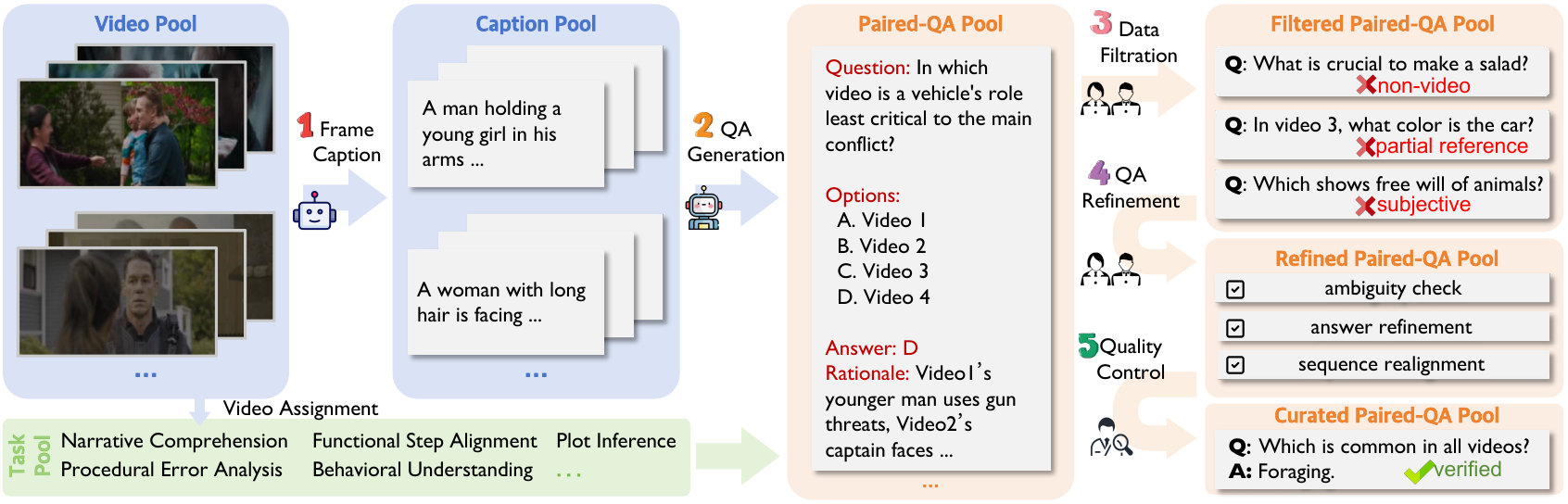}
\caption{Illustration of the CrossVid annotation pipeline. The process consists of the following main stages: (1) Frames are extracted from videos and captioned by Qwen2.5-VL-72B; (2) Deepseek-R1 generates QA pairs using task-specific prompts; (3) The QA pairs undergo rigorous human quality review, including data filtering, refinement, and quality control.}
\label{fig}
\end{figure*}

We design a semi-automated multi-stage pipeline to construct CrossVid.
The overall process is shown in Figure~\ref{fig}.

\noindent\textbf{Frame Caption}
We first densely extract frames from the source video and leverage Qwen2.5-VL-72B~\cite{qwen2.5} to generate concise captions for each extracted frame.
In order to enrich the contextual information for the caption, we also incorporate metadata from the original datasets (\emph{e.g.}, plot summaries, scene descriptions, and action labels) during the generation process.

\noindent\textbf{QA Generation}
Firstly, we manually assign the most suitable videos to the predefined tasks. 
For example, cooking videos inherently comprise multiple sequential steps, making them appropriate for temporal understanding tasks. 
More details about the assignment process can be found in the Appendix.
Subsequently, for each task, videos are clustered into different sets based on their labels in the original datasets. 
Videos sharing the same labels (\emph{e.g.
}, the same film genre in MovieChat-1K or the same recipe in YouCook2) are grouped together. 
We then randomly sample from the same set the number of videos required for each question, and provide their frame-level captions to DeepSeek-R1~\cite{deepseek-r1} for automatic QA generation.
We strictly retrieve videos from the same set to ensure strong semantic relevance and comparability across videos.
For each task, we design a customized prompt comprising three key components:
1) The prompt explicitly instructs DeepSeek-R1 to analyze the relationships among all given videos. 
2) The prompt guides DeepSeek-R1 to generate QA pairs that are closely aligned with the specific requirements of the task (\emph{e.g.}, behavioral understanding tasks may prioritize comparison of action patterns and aims). 
3) The prompt requires DeepSeek-R1 to provide a detailed explanation to support the correctness of its answer. 
Such prompts reduce DeepSeek-R1's hallucinations during the QA generation process and enhance the reliability of the generated QA pairs.
Furthermore, this ensures that the generated QA pairs are challenging and require performing integrative reasoning across multiple videos.

\noindent\textbf{Data Filtration}
To ensure data quality, we conduct a rigorous manual review with ten expert annotators. 
During this coarse filtering stage, we sequentially eliminate unsuitable QA pairs through three steps.
First, we filter out questions unrelated to video understanding.
Then, we exclude questions referencing only specific queried videos (\emph{e.g.}, ``In video three, what color is the car?"). 
Finally, we discard subjective or overly complex questions, such as those requiring philosophical reasoning or domain-specific expertise.

\noindent\textbf{QA Refinement}
The retained QA pairs then undergo refinement consisting of three steps:
1) Annotators revise the questions to eliminate ambiguities.
2) Each annotator then answers questions without consulting DeepSeek R1's outputs or explanations.
3) Based on the annotators' responses, task-specific refinements are further conducted.
Specifically, for single- and multi-choice questions, both ground truths and other false options are refined to ensure unique correctness.
For the functional step sequencing task in temporal understanding, we address potential shortcut learning (reliance on camera angle continuity) by temporal realignment, \emph{i.e.}, each preceding clip is advanced by 1-5 seconds while equivalent offsets delay subsequent clips.
Such a strategy creates intentional discontinuity in visual features across clip boundaries, forcing models to infer temporal relationships through semantic content rather than low-level consistency.
For open-ended questions, annotators check whether the scoring points align with the generated standard answer and cover all key information related to the question.

\subsubsection{Quality Control}
At the quality control step, an independent group of experts further assesses the refined paired QA pool and forms the curated pool.
These processes are conducted via our designed interface, shown in the Appendix.

Through this semi-automated pipeline, a large number of high-quality QA pairs are generated. More importantly, our curation process ensures that each QA pair is built on meaningful inter-video relationships and requires integrative CVR. This meets the objectives of our CrossVid.

\section{Experiments and Analysis}
We conduct a comprehensive evaluation of existing MLLMs on CrossVid. In this section, we first describe experimental settings, followed by a detailed analysis of model performance. We then present ablation studies and key findings.

\subsection{Experimental Settings}
We evaluate 22 MLLMs on CrossVid, including both closed-source models (\emph{e.g.}, GPT-4.1 and Gemini-2.5-Pro) and a wide range of open-source models (\emph{e.g.}, Qwen2.5-VL, InternVL3) with parameter sizes spanning from 7B to 78B. Other architectures, like Mixture of Experts (MoE), are also included.
For video pre-processing, we evenly distribute the total number of frames across all input videos and uniformly sample frames within each video.
For each QA pair, the frames of all videos and the question prompt are simultaneously input into the MLLM in one turn.
We adopt a zero-shot strategy and require the MLLMs to give their answers directly.
Inference on open-source models is conducted according to their official implementations, whereas closed-source models are accessed via their official APIs. 
We report accuracy as our evaluation metric. 
More details of the implementation and evaluation can be found in the Appendix.

\begin{table*}[!ht]
\centering
\setlength{\tabcolsep}{1mm}
{
{\small
\begin{tabular}{lcccccccccccccc}
\toprule[1pt]
\multirow{2}{*}{\textbf{Models}} & \multirow{2}{*}{\textbf{O.Avg}} & \multicolumn{5}{c}{\textbf{Comparative Analysis}} & \multicolumn{4}{c}{\textbf{Temporal Understanding}} & \multicolumn{3}{c}{\textbf{Multi-view Reasoning}} & \textbf{Free-form} \\ \cmidrule(lr){3-7}  \cmidrule(lr){8-11}  \cmidrule(lr){12-14}  \cmidrule(lr){15-15} 
                        &                                              & BU     & NC     & CC    & PEA   & \textbf{C.Avg}  & PI       & FSA      & PSS      & \textbf{T.Avg}     & MSR         & MOC         & \textbf{M.Avg}        & CCQA         \\ \midrule 
Human                    &   89.2        & 85.6   & 92.3   & 90.7  & 83.9  & 88.1   & 91.6     & 85.2     & 89.9     & 88.9      & 93.2        & 94.2        & 93.7         & 85.2         \\ \midrule
\rowcolor{gray!10}
\multicolumn{15}{l}{\textit{Closed-Source Models}}                                                                                                                                                                                              \\
GPT-4.1~(2025)                                       & 45.2                         & 46.2   & 34.6   & 58.5  & \textbf{51.2}  & 47.6   & 70.9     & 8.6      & 60.5     & 46.7      & \textbf{38.6}        & 38.2        & 38.4         & 44.6         \\
GPT-4o~\shortcite{gpt4o}                                        & 36.8                         & 38.2   & 34.3   & 50.7  & 49.1  & 43.1   & 57.8     & 9.1      & 39.7     & 35.5      & 15.3        & 39.4        & 27.4         & 34.2         \\
Doubao-1.5-VL-pro ~\shortcite{doubao}                      & 44.3                         & 51.2   & \textbf{58.1}   & \textbf{69.5}  & 36.4  & 53.8   & 66.9     & 4.6      & 36.8     & 36.1      & 37.4        & 32.0          & 34.7         & 50.1         \\
Gemini-2.5-Pro    ~\shortcite{gemini-25-pro}                         & \textbf{50.4}                & \textbf{54.2} & 51.8   & 68.7  & 44.1  & \textbf{54.7}   & 76.5     & \textbf{13.4}     & \textbf{78.2}     & \textbf{56.0}      & 32.0        & 25.3        & 28.7         & \textbf{59.8}         \\ \midrule
\rowcolor{gray!10}
\multicolumn{15}{l}{\textit{Open-Source Models $\sim$ MoE}}                                                                                                                                                                                    \\
Kimi-VL-A3B-Thinking ~\shortcite{kimi}                        & 28.2                         & 29.4   & 33.3   & 36.8  & 34.0  & 33.4   & 40.6     & 3.8      & 9.2      & 17.9      & 28.4        & 36.9        & 32.7         & 29.2         \\
ERNIE-4.5-VL-A3B ~\shortcite{ernie}                        & 24.8                         & 12.6   & 28.2   & 24.2  & 36.4  & 25.4   & 52.6     & 4.0      & 2.4      & 19.7      & 29.6        & 35.3        & 32.5         & 22.5         \\ \midrule
\rowcolor{gray!10}
\multicolumn{15}{l}{\textit{Open-Source Models \textless 10B}}                                                                                                                                                                                \\
Qwen2.5-VL-7B ~\shortcite{qwen2.5}                            & 18.3                         & 19.6   & 19.0   & 23.4  & 15.0  & 19.3   & 58.6     & 1.2      & 0.3      & 20.0      & 11.8        & 21.7        & 16.8         & 12.0           \\
InternVL3-8B ~\shortcite{internvl3}                              & 25.6                         & 15.2   & 22.8   & 24.3  & 42.1  & 26.1   & 56.2     & 3.2      & 1.5      & 20.3      & 34.0        & \textbf{47.3}        & \textbf{40.7}         & 9.7          \\
LongVA-7B-DPO ~\shortcite{longva}                            & 18.0                         & 16.2   & 20.6   & 18.2  & 39.0  & 23.5   & 18.7     & 2.1      & 1.8      & 7.5       & 24.2        & 28.4        & 26.3         & 10.7         \\
VideoLLaMA3-7B ~\shortcite{videollama3}                            & 15.3                         & 14.7   & 19.5   & 22.2  & 26.6  & 20.8   & 11.6     & 5.1      & 3.5      & 6.7       & 18.7        & 20.8        & 19.8         & 9.8          \\
Qwen2.5-Omni-7B ~\shortcite{qwen2.5-omni}                           & 24.6                         & 27.5   & 26.0   & 32.7  & 20.4  & 26.7   & 60.2     & 0.4      & 4.1      & 21.6      & 23.2        & 36.0        & 29.6         & 15.3         \\
Phi-3.5-vision ~\shortcite{phi-35-vision}                            & 21.5                         & 18.3   & 22.0   & 21.8  & 41.5  & 25.9   & 46.2     & 1.2      & 4.1      & 17.2      & 28.4        & 26.7        & 27.6         & 4.3          \\
MiniCPM-O\ 2.6 ~\shortcite{minicpm}                           & 25.6                         & 20.3   & 21.8   & 20.1  & 42.6  & 26.2   & 72.1     & 2.9      & 4.1      & 26.4      & 27.1        & 35.7        & 31.4         & 9.0          \\
MiMo-7B ~\shortcite{mimo}                                    & 28.3                         & 22.3   & 30.6   & 39.2  & 32.8  & 31.2   & 54.6     & 2.8      & 11.6     & 23.0      & 25.8        & 41.3        & 33.6         & 22.0         \\
Video-R1-7B ~\shortcite{videor1}                                & 21.6                         & 14.7   & 23.0   & 19.9  & 16.3  & 18.5   & \textbf{77.3}     & 1.9      & 1.5      & 26.9      & 19.4        & 34.4        & 26.9         & 8.0          \\
GLM-4.1V-9B-Thinking ~\shortcite{glm-4.1}                      & 35.1                         & 49.8   & 39.9   & 50.6  & 38.6  & 44.7   & 50.2     & 5.1      & 14.1     & 23.1      & 36.7        & 38.9        & 37.8         & 26.9         \\ \midrule
\rowcolor{gray!10}
\multicolumn{15}{l}{\textit{Open-Source Models $\sim$30B}}                                                                                                                                                                                    \\
Qwen2.5-VL-32B ~\shortcite{qwen2.5}                             & 33.7                         & 31.4   & 39.5   & 48.6  & 33.7  & 38.3   & 65.7     & 5.1      & 8.7      & 26.5      & 23.7        & 39.6        & 31.7         & 41.2         \\
InternVL3-38B ~\shortcite{internvl3}                            & 23.5                         & 15.9   & 33.7   & 33.6  & 27.9  & 27.8   & 24.3     & 4.5      & 1.5      & 10.1      & 40.4        & 36.8        & 38.6         & 16.2         \\ \midrule
\rowcolor{gray!10}
\multicolumn{15}{l}{\textit{Open-Source Models $\sim$70B}}                                                                                                                                                                                    \\
Qwen2.5-VL-72B ~\shortcite{qwen2.5}                             & 34.4                         & 37.7   & 38.5   & 52.6  & 39.6  & 42.1   & 61.8     & 5.9      & 20.0       & 29.2      & 20.9        & 26.0        & 23.5         & 41.2         \\
InternVL3-78B ~\shortcite{internvl3}                           & 25.8                         & 27.1   & 29.4   & 33.1  & 42.7  & 33.1   & 37.5     & 4.9      & 4.4      & 15.6      & 22.9        & 33.2        & 28.1         & 23.2         \\
LLaVA-Video-72B~\shortcite{llava-video}                     & 27.5                         & 26.5   & 26.5   & 34.0  & 48.7  & 33.9   & 51.8     & 3.2      & 11.1     & 22.0        & 26.8        & 29.0        & 27.9         & 17.8         \\
LLaVA-OV-72B~\shortcite{llava-ov}                         & 27.5                         & 22.1   & 23.3   & 29.2  & 37.0  & 27.9   & 74.1     & 8.8      & 5.0      & 29.3      & 25.9        & 35.0        & 30.5         & 14.6         \\ 
\bottomrule[1pt]
\end{tabular}
}
}
\small\caption{Performance of 22 evaluated MLLMs on CrossVid. \textbf{O.Avg}: the average accuracy of ten tasks. \textbf{C.Avg}: the average accuracy of comparative analysis; \textbf{T.Avg}: the average accuracy of temporal understanding; \textbf{M.Avg}: the average accuracy of multi-view reasoning. The top result in each task is highlighted in bold.}
\label{table_main_results}
\end{table*}

\subsection{Main Results}
We present the performance of MLLMs on CrossVid, including the accuracy for each task, the average accuracy of each dimension, and the overall average on all tasks, in Table~\ref{table_main_results}. 
Based on these, we highlight three main observations:

\noindent \textbf{1) CVR is challenging for existing MLLMs.}
The average performance of all MLLMs is notably lower than the human performance of 89.2\%. 
Even the best-performing MLLM, Gemini-2.5-Pro, only achieves an overall average accuracy of 50.4\%.
Notably, MLLMs perform worse than humans on multi-view reasoning, a task focusing on spatial reasoning.
Specifically, the leading MLLM, InternVL3-8B, only achieves an average accuracy of  40.7\%, while humans reach 93.7\%. 
The gap between MLLM and humans becomes even larger in temporal understanding.
For example, on the step alignment task, Gemini-2.5-Pro with the highest accuracy achieves only 13.4\% versus 85.2\% for humans. 
These reveal critical limitations in existing MLLMs' capability for both temporal and spatial understanding in CVR.

\noindent \textbf{2) Closed-source MLLMs substantially outperform open-source counterparts.}
All closed-source MLLMs obtain higher overall average accuracies than open-source MLLMs, and the advantage of closed-source MLLMs is much more prominent on several crucial tasks. 
Notably, for temporal understanding, closed-source MLLMs consistently outperform open-source MLLMs. 
The closed-source GPT-4o with the lowest score achieves an average accuracy of 35.5\%, which is still 6.2\% higher than the leading open-source model LLaVA-OV. 

\noindent \textbf{3) ``Thinking" -enabled models demonstrate performance gains.}
Among closed-source models, those featuring explicit reasoning modules (\emph{e.g.}, Gemini-2.5-Pro) consistently achieve the highest overall and per-task accuracies. 
In the 10B parameter group, the top two models are thinking-enabled GLM-4.1V-9B-Thinking (35.1\%) and MiMo-7B (28.3\%), which outperform the third best by margins of 9.5\% and 2.7\%, respectively.
Therefore, internal “thinking” mechanism allows models to better structure multi-step reasoning processes, which contributes to enhanced performance on complex cross-video tasks.

\subsection{Ablation Study}
We conduct ablations to better understand MLLMs' performance on CVR and present more insights.

\subsection{Impact of frame number}
The number of input frames determines the amount of visual information available to the model for reasoning. To assess its impact, we evaluate Qwen2.5-VL-72B on CrossVid using 32, 64, 128, and 256 total input frames. 
For settings with fewer than 256 frames, frames are uniformly sampled from the full set of 256 frames.
Results are reported in Table~\ref{table_frame}.

It can be observed that increasing the number of input frames generally improves model performance. The improvements are most pronounced in tasks that require comprehensive context, such as comparative analysis and free-form QA. For Qwen2.5-VL-72B, the overall accuracy increases by 5.7\% (from 33.8\% to 39.5\%) as frame count rises from 32 to 256, with even larger improvements (up to 15.1\%) observed on the open-ended CCQA task.

With more frames, the model can access richer visual information, which is crucial for answering cross-video questions requiring precise details.
For instance, when answering an open-ended question to tell the difference between cooking methods in two videos, the model can only identify superficial actions at 32 frames. When increasing the frame count to 64, the model is able to distinguish specific core techniques. At 256 frames, its analysis became granular enough to recognize secondary ingredients.

However, excessive irrelevant frames may introduce noise, which leads to information redundancy and distracts the model with irrelevant content. For example, when solving a plot inference problem with the beginning and ending of a war film, the model correctly identifies the key events (\emph{e.g.}, a troop convoy and a negotiation scene) with lower frame counts like 32 and 64. However, as more frames are provided, irrelevant atmospheric information (\emph{e.g.}, generic shots of injured soldiers) distracts the model from the primary causal chain, leading to the incorrect answer based on broad military planning associations.

These findings provide guidance for future advancements on CVR. On one hand, expanding the model's context window allows it to perceive more information; On the other hand, key frame selection helps to filter out irrelevant clues and make models concentrate on core information.

\begin{table}[!htbp]
\centering
{\small
\begin{tabular}{cccccc}
\toprule[1pt]
\textbf{\#Frames} & \textbf{O.Avg} & \textbf{C.Avg} & \textbf{T.Avg} & \textbf{M.Avg} & \textbf{CCQA} \\
\midrule
32      &       33.8&       37.0&       33.8&       35.1&      18.9\\
64      &       36.9&       39.8&       37.4&       35.9&      25.9\\
128     &       39.1&       45.7&       \textbf{34.5}&       \textbf{36.4}&      32.0\\
256     &       \textbf{39.5}&       \textbf{47.5}&       33.9&       34.9&      \textbf{34.0}\\ 
\bottomrule[1pt]
\end{tabular}
}
\small\caption{Comparison results of performance under different numbers of input frames.}
\label{table_frame}
\end{table}

\subsection{Effectiveness of CoT prompts}
Previous results show that internally thinking-enabled MLLMs outperform their counterparts. 
To assess the effectiveness of Chain-of-Thought (CoT) on non-thinking MLLMs, we design prompts to explicitly require a three-stage process: 
1) comprehend the question, 
2) analyze the frames for each input video, and 
3) aggregate information from all videos before answering. 
Details of prompts can be found in the Appendix.
MLLMs are required to output each step of their reasoning process.
We conduct experiments on GPT-4.1 and three other open-source MLLMs in different parameter size groups.
We keep the frames for each MLLM the same as the previous direct answering strategy.

Table~\ref{table_cot} shows the performance comparison with and without CoT prompting. For most MLLMs, CoT brings performance gains on temporal understanding and multi-view reasoning tasks. 
This indicates that CoT facilitates more systematic reasoning across videos on both temporal and spatial understanding tasks.
Notably, CoT prompting does not consistently improve accuracy on each task, whereas open-source MLLMs with more parameters exhibit larger overall performance gains. This suggests that larger MLLMs are more capable of benefiting from prompt-based optimization.

\begin{table}[htbp]
\centering
{\small
\setlength{\tabcolsep}{1.3mm}
\begin{tabular}{lccccc}
\toprule[1pt]
\textbf{Method} &    \textbf{O.Avg} & \textbf{C.Avg} & \textbf{T.Avg} & \textbf{M.Avg} & \textbf{CCQA} \\ \midrule
\rowcolor{gray!10}
GPT-4.1        &       &       &       &       &      \\
~w/o CoT       & \textbf{45.2}  & \textbf{47.6}  & 46.7  & 38.4  & \textbf{44.6} \\
~w/ CoT      &       44.9&       46.7&       \textbf{48.2}&       \textbf{40.4}&      36.7\\ \midrule
\rowcolor{gray!10}
MiniCPM-o 2.6 &       &       &       &       &      \\
~w/o CoT   & \textbf{25.6}  & 26.2  & \textbf{26.4}  & 31.4  & \textbf{9.0}    \\
~w/ CoT       &       23.7&       \textbf{26.7}&       18.7&       \textbf{33.3}&      7.2\\ \midrule
\rowcolor{gray!10}
InternVL3-38B  &       &       &       &       &      \\
~w/o CoT      & 23.5  & \textbf{27.8}  & 10.1  & \textbf{38.6}  & 16.2 \\
~w/ CoT       &       \textbf{24.4}&       26.3&       \textbf{16.7}&       35.2&      \textbf{18.0}\\ \midrule
\rowcolor{gray!10}
Qwen2.5-VL-72B &       &       &       &       &      \\
~w/o CoT      & 34.4  & 42.1  & 29.2  & 23.5  & \textbf{41.2} \\
~w/ CoT         &       \textbf{39.5}&       \textbf{47.5}&       \textbf{33.9}&       \textbf{34.9}&      34.0\\ 
\bottomrule[1pt]
\end{tabular}
}
\small\caption{Comparison of performances with and without CoT prompting.}
\label{table_cot}
\end{table}

\subsection{Error Analysis}
To further examine the CVR capabilities of current MLLMs and better understand their limitations, we manually analyze the errors in their reasoning steps and identify four primary error types. The percentage and detailed examples for each error are presented in the Appendix.

\textbf{(a) Key frame loss:}
Compared with previous single-video understanding, our tasks require multiple videos to be input simultaneously, which reduces the number of frames for each video. This may result in the loss of core information. As a result, MLLMs may fail to obtain the necessary information to answer the question and thus provide incorrect answers.

\textbf{(b) Video understanding error:}
In this type of error, although MLLMs capture the key information from each video, they still might fall short in cross-video understanding. 
Since their analysis of individual videos might be insufficient due to the requirement of simultaneously processing multiple videos, 
the failure to understand any single video leads to errors in multi-video understanding as a whole.

\textbf{(c) Cross-video comparison error:}
Although MLLMs are capable of correctly understanding each individual video, they may still struggle with cross-video comparison. 

For example, when the MLLM is asked to identify the hug in which film represents the crisis resolution, the MLLM successfully identifies the hugs in all videos in the group but fails to reason and compare their meanings in the context.

\textbf{(d) Format error:}
CrossVid contains tasks requiring specific output formats, \emph{e.g.}, a time interval with both beginning and ending timestamps in the functional step alignment task. 
However, some MLLMs might fail to accurately follow the specific instructions or constraints described in the prompt, resulting in the failure of answer extraction.

\section{Conclusion}
We present CrossVid, the first comprehensive benchmark for evaluating MLLMs on CVR. CrossVid is constructed through a semi-automated pipeline and strict multi-stage quality control, resulting in 10 diverse tasks spanning 4 key reasoning dimensions. Our experimental evaluation of 22 frontier MLLMs reveals significant challenges in CVR, with the best model (Gemini-2.5-Pro) achieving only moderate performance far below human levels. Extensive ablation studies and error analyses provide deep insights into the limitations of current models.
We hope CrossVid will serve as a valuable resource to drive advances in multi-video understanding and robust, generalizable visual reasoning.

\bibliography{aaai2026}

@article{qwen2.5,
  title={Qwen2. 5-vl technical report},
  author={Bai, Shuai and Chen, Keqin and Liu, Xuejing and Wang, Jialin and Ge, Wenbin and Song, Sibo and Dang, Kai and Wang, Peng and Wang, Shijie and Tang, Jun and others},
  journal={arXiv preprint arXiv:2502.13923},
  year={2025}
}

@article{internvl3,
  title={Internvl3: Exploring advanced training and test-time recipes for open-source multimodal models},
  author={Zhu, Jinguo and Wang, Weiyun and Chen, Zhe and Liu, Zhaoyang and Ye, Shenglong and Gu, Lixin and Tian, Hao and Duan, Yuchen and Su, Weijie and Shao, Jie and others},
  journal={arXiv preprint arXiv:2504.10479},
  year={2025}
}

@article{video-llava,
  title={Video-llava: Learning united visual representation by alignment before projection},
  author={Lin, Bin and Ye, Yang and Zhu, Bin and Cui, Jiaxi and Ning, Munan and Jin, Peng and Yuan, Li},
  journal={arXiv preprint arXiv:2311.10122},
  year={2023}
}

@article{sharegpt4video,
  title={Sharegpt4video: Improving video understanding and generation with better captions},
  author={Chen, Lin and Wei, Xilin and Li, Jinsong and Dong, Xiaoyi and Zhang, Pan and Zang, Yuhang and Chen, Zehui and Duan, Haodong and Tang, Zhenyu and Yuan, Li and others},
  journal={Advances in Neural Information Processing Systems},
  volume={37},
  pages={19472--19495},
  year={2024}
}

@article{videollama,
  title={Video-llama: An instruction-tuned audio-visual language model for video understanding},
  author={Zhang, Hang and Li, Xin and Bing, Lidong},
  journal={arXiv preprint arXiv:2306.02858},
  year={2023}
}

@inproceedings{videomme,
  title={Video-mme: The first-ever comprehensive evaluation benchmark of multi-modal llms in video analysis},
  author={Fu, Chaoyou and Dai, Yuhan and Luo, Yongdong and Li, Lei and Ren, Shuhuai and Zhang, Renrui and Wang, Zihan and Zhou, Chenyu and Shen, Yunhang and Zhang, Mengdan and others},
  booktitle={Proceedings of the Computer Vision and Pattern Recognition Conference},
  pages={24108--24118},
  year={2025}
}

@inproceedings{nextqa,
  title={Next-qa: Next phase of question-answering to explaining temporal actions},
  author={Xiao, Junbin and Shang, Xindi and Yao, Angela and Chua, Tat-Seng},
  booktitle={Proceedings of the IEEE/CVF conference on computer vision and pattern recognition},
  pages={9777--9786},
  year={2021}
}

@article{all-angels-bench,
  title={Seeing from another perspective: Evaluating multi-view understanding in mllms},
  author={Yeh, Chun-Hsiao and Wang, Chenyu and Tong, Shengbang and Cheng, Ta-Ying and Wang, Ruoyu and Chu, Tianzhe and Zhai, Yuexiang and Chen, Yubei and Gao, Shenghua and Ma, Yi},
  journal={arXiv preprint arXiv:2504.15280},
  year={2025}
}

@inproceedings{key-frame1,
  title={Adaptive keyframe sampling for long video understanding},
  author={Tang, Xi and Qiu, Jihao and Xie, Lingxi and Tian, Yunjie and Jiao, Jianbin and Ye, Qixiang},
  booktitle={Proceedings of the Computer Vision and Pattern Recognition Conference},
  pages={29118--29128},
  year={2025}
}

@inproceedings{key-frame2,
  title={The devil is in temporal token: High quality video reasoning segmentation},
  author={Gong, Sitong and Zhuge, Yunzhi and Zhang, Lu and Yang, Zongxin and Zhang, Pingping and Lu, Huchuan},
  booktitle={Proceedings of the Computer Vision and Pattern Recognition Conference},
  pages={29183--29192},
  year={2025}
}

@inproceedings{video-xl,
  title={Video-xl: Extra-long vision language model for hour-scale video understanding},
  author={Shu, Yan and Liu, Zheng and Zhang, Peitian and Qin, Minghao and Zhou, Junjie and Liang, Zhengyang and Huang, Tiejun and Zhao, Bo},
  booktitle={Proceedings of the Computer Vision and Pattern Recognition Conference},
  pages={26160--26169},
  year={2025}
}

@article{reilly2025my,
  title={From My View to Yours: Ego-Augmented Learning in Large Vision Language Models for Understanding Exocentric Daily Living Activities},
  author={Reilly, Dominick and Govind, Manish Kumar and Xue, Le and Das, Srijan},
  journal={arXiv preprint arXiv:2501.05711},
  year={2025}
}

@article{deepseek-r1,
  title={Deepseek-r1: Incentivizing reasoning capability in llms via reinforcement learning},
  author={Guo, Daya and Yang, Dejian and Zhang, Haowei and Song, Junxiao and Zhang, Ruoyu and Xu, Runxin and Zhu, Qihao and Ma, Shirong and Wang, Peiyi and Bi, Xiao and others},
  journal={arXiv preprint arXiv:2501.12948},
  year={2025}
}

@article{videor1,
  title={Video-r1: Reinforcing video reasoning in mllms},
  author={Feng, Kaituo and Gong, Kaixiong and Li, Bohao and Guo, Zonghao and Wang, Yibing and Peng, Tianshuo and Wu, Junfei and Zhang, Xiaoying and Wang, Benyou and Yue, Xiangyu},
  journal={arXiv preprint arXiv:2503.21776},
  year={2025}
}

@article{glm-4.1,
  title={GLM-4.1 V-Thinking: Towards Versatile Multimodal Reasoning with Scalable Reinforcement Learning},
  author={Hong, Wenyi and Yu, Wenmeng and Gu, Xiaotao and Wang, Guo and Gan, Guobing and Tang, Haomiao and Cheng, Jiale and Qi, Ji and Ji, Junhui and Pan, Lihang and others},
  journal={arXiv preprint arXiv:2507.01006},
  year={2025}
}

@inproceedings{mvbench,
  title={Mvbench: A comprehensive multi-modal video understanding benchmark},
  author={Li, Kunchang and Wang, Yali and He, Yinan and Li, Yizhuo and Wang, Yi and Liu, Yi and Wang, Zun and Xu, Jilan and Chen, Guo and Luo, Ping and others},
  booktitle={Proceedings of the IEEE/CVF Conference on Computer Vision and Pattern Recognition},
  pages={22195--22206},
  year={2024}
}

@inproceedings{activitynet,
  title={Activitynet-qa: A dataset for understanding complex web videos via question answering},
  author={Yu, Zhou and Xu, Dejing and Yu, Jun and Yu, Ting and Zhao, Zhou and Zhuang, Yueting and Tao, Dacheng},
  booktitle={Proceedings of the AAAI Conference on Artificial Intelligence},
  volume={33},
  pages={9127--9134},
  year={2019}
}

@inproceedings{ego-exo4d,
  title={Ego-exo4d: Understanding skilled human activity from first-and third-person perspectives},
  author={Grauman, Kristen and Westbury, Andrew and Torresani, Lorenzo and Kitani, Kris and Malik, Jitendra and Afouras, Triantafyllos and Ashutosh, Kumar and Baiyya, Vijay and Bansal, Siddhant and Boote, Bikram and others},
  booktitle={Proceedings of the IEEE/CVF Conference on Computer Vision and Pattern Recognition},
  pages={19383--19400},
  year={2024}
}

@article{tvqa,
  title={Tvqa: Localized, compositional video question answering},
  author={Lei, Jie and Yu, Licheng and Bansal, Mohit and Berg, Tamara L},
  journal={arXiv preprint arXiv:1809.01696},
  year={2018}
}

@article{longvideobench,
  title={Longvideobench: A benchmark for long-context interleaved video-language understanding},
  author={Wu, Haoning and Li, Dongxu and Chen, Bei and Li, Junnan},
  journal={Advances in Neural Information Processing Systems},
  volume={37},
  pages={28828--28857},
  year={2024}
}

@inproceedings{mmvu,
  title={Mmvu: Measuring expert-level multi-discipline video understanding},
  author={Zhao, Yilun and Zhang, Haowei and Xie, Lujing and Hu, Tongyan and Gan, Guo and Long, Yitao and Hu, Zhiyuan and Chen, Weiyuan and Li, Chuhan and Xu, Zhijian and others},
  booktitle={Proceedings of the Computer Vision and Pattern Recognition Conference},
  pages={8475--8489},
  year={2025}
}

@inproceedings{egoexolearn,
  title={Egoexolearn: A dataset for bridging asynchronous ego-and exo-centric view of procedural activities in real world},
  author={Huang, Yifei and Chen, Guo and Xu, Jilan and Zhang, Mingfang and Yang, Lijin and Pei, Baoqi and Zhang, Hongjie and Dong, Lu and Wang, Yali and Wang, Limin and others},
  booktitle={Proceedings of the IEEE/CVF Conference on Computer Vision and Pattern Recognition},
  pages={22072--22086},
  year={2024}
}

@article{mlvu,
  title={Mlvu: A comprehensive benchmark for multi-task long video understanding},
  author={Zhou, Junjie and Shu, Yan and Zhao, Bo and Wu, Boya and Xiao, Shitao and Yang, Xi and Xiong, Yongping and Zhang, Bo and Huang, Tiejun and Liu, Zheng},
  journal={arXiv e-prints},
  pages={arXiv--2406},
  year={2024}
}

@inproceedings{animalkingdom,
  title={Animal kingdom: A large and diverse dataset for animal behavior understanding},
  author={Ng, Xun Long and Ong, Kian Eng and Zheng, Qichen and Ni, Yun and Yeo, Si Yong and Liu, Jun},
  booktitle={Proceedings of the IEEE/CVF conference on computer vision and pattern recognition},
  pages={19023--19034},
  year={2022}
}

@inproceedings{moviechat,
  title={Moviechat: From dense token to sparse memory for long video understanding},
  author={Song, Enxin and Chai, Wenhao and Wang, Guanhong and Zhang, Yucheng and Zhou, Haoyang and Wu, Feiyang and Chi, Haozhe and Guo, Xun and Ye, Tian and Zhang, Yanting and others},
  booktitle={Proceedings of the IEEE/CVF Conference on Computer Vision and Pattern Recognition},
  pages={18221--18232},
  year={2024}
}

@inproceedings{youcook,
  title={Towards automatic learning of procedures from web instructional videos},
  author={Zhou, Luowei and Xu, Chenliang and Corso, Jason},
  booktitle={Proceedings of the AAAI conference on artificial intelligence},
  volume={32},
  year={2018}
}

@article{visdrone,
  title={Robust multi-drone multi-target tracking to resolve target occlusion: A benchmark},
  author={Liu, Zhihao and Shang, Yuanyuan and Li, Timing and Chen, Guanlin and Wang, Yu and Hu, Qinghua and Zhu, Pengfei},
  journal={IEEE Transactions on Multimedia},
  volume={25},
  pages={1462--1476},
  year={2023},
  publisher={IEEE}
}

@article{charades,
    author = {Gunnar A. Sigurdsson and G{\"u}l Varol and Xiaolong Wang and Ivan Laptev and Ali Farhadi and Abhinav Gupta},
    title = {Hollywood in Homes: Crowdsourcing Data Collection for Activity Understanding},
    journal = {ArXiv e-prints},
    eprint = {1604.01753}, 
    year = {2016},
    url = {http://arxiv.org/abs/1604.01753},
}

@inproceedings{assembly101,
  title={Assembly101: A large-scale multi-view video dataset for understanding procedural activities},
  author={Sener, Fadime and Chatterjee, Dibyadip and Shelepov, Daniel and He, Kun and Singhania, Dipika and Wang, Robert and Yao, Angela},
  booktitle={Proceedings of the IEEE/CVF Conference on Computer Vision and Pattern Recognition},
  pages={21096--21106},
  year={2022}
}

@article{doubao,
  title={Seed1. 5-vl technical report},
  author={Guo, Dong and Wu, Faming and Zhu, Feida and Leng, Fuxing and Shi, Guang and Chen, Haobin and Fan, Haoqi and Wang, Jian and Jiang, Jianyu and Wang, Jiawei and others},
  journal={arXiv preprint arXiv:2505.07062},
  year={2025}
}

@article{kimi,
  title={Kimi-vl technical report},
  author={Team, Kimi and Du, Angang and Yin, Bohong and Xing, Bowei and Qu, Bowen and Wang, Bowen and Chen, Cheng and Zhang, Chenlin and Du, Chenzhuang and Wei, Chu and others},
  journal={arXiv preprint arXiv:2504.07491},
  year={2025}
}

@article{longva,
  title={Long context transfer from language to vision},
  author={Zhang, Peiyuan and Zhang, Kaichen and Li, Bo and Zeng, Guangtao and Yang, Jingkang and Zhang, Yuanhan and Wang, Ziyue and Tan, Haoran and Li, Chunyuan and Liu, Ziwei},
  journal={arXiv preprint arXiv:2406.16852},
  year={2024}
}

@article{videollama3,
  title={Videollama 3: Frontier multimodal foundation models for image and video understanding},
  author={Zhang, Boqiang and Li, Kehan and Cheng, Zesen and Hu, Zhiqiang and Yuan, Yuqian and Chen, Guanzheng and Leng, Sicong and Jiang, Yuming and Zhang, Hang and Li, Xin and others},
  journal={arXiv preprint arXiv:2501.13106},
  year={2025}
}

@article{qwen2.5-omni,
  title={Qwen2. 5-omni technical report},
  author={Xu, Jin and Guo, Zhifang and He, Jinzheng and Hu, Hangrui and He, Ting and Bai, Shuai and Chen, Keqin and Wang, Jialin and Fan, Yang and Dang, Kai and others},
  journal={arXiv preprint arXiv:2503.20215},
  year={2025}
}

@article{minicpm,
  title={Minicpm-v: A gpt-4v level mllm on your phone},
  author={Yao, Yuan and Yu, Tianyu and Zhang, Ao and Wang, Chongyi and Cui, Junbo and Zhu, Hongji and Cai, Tianchi and Li, Haoyu and Zhao, Weilin and He, Zhihui and others},
  journal={arXiv preprint arXiv:2408.01800},
  year={2024}
}

@article{mimo,
  title={MiMo: Unlocking the Reasoning Potential of Language Model--From Pretraining to Posttraining},
  author={Xiaomi, LLM and Xia, Bingquan and Shen, Bowen and Zhu, Dawei and Zhang, Di and Wang, Gang and Zhang, Hailin and Liu, Huaqiu and Xiao, Jiebao and Dong, Jinhao and others},
  journal={arXiv preprint arXiv:2505.07608},
  year={2025}
}

@article{llava-video,
  title={Video instruction tuning with synthetic data},
  author={Zhang, Yuanhan and Wu, Jinming and Li, Wei and Li, Bo and Ma, Zejun and Liu, Ziwei and Li, Chunyuan},
  journal={arXiv preprint arXiv:2410.02713},
  year={2024}
}

@article{llava-ov,
  title={Llava-onevision: Easy visual task transfer},
  author={Li, Bo and Zhang, Yuanhan and Guo, Dong and Zhang, Renrui and Li, Feng and Zhang, Hao and Zhang, Kaichen and Zhang, Peiyuan and Li, Yanwei and Liu, Ziwei and others},
  journal={arXiv preprint arXiv:2408.03326},
  year={2024}
}

@article{gpt4o,
  title={Gpt-4o system card},
  author={Hurst, Aaron and Lerer, Adam and Goucher, Adam P and Perelman, Adam and Ramesh, Aditya and Clark, Aidan and Ostrow, AJ and Welihinda, Akila and Hayes, Alan and Radford, Alec and others},
  journal={arXiv preprint arXiv:2410.21276},
  year={2024}
}

@article{gemini-25-pro,
  title={Gemini 2.5: Pushing the frontier with advanced reasoning, multimodality, long context, and next generation agentic capabilities},
  author={Comanici, Gheorghe and Bieber, Eric and Schaekermann, Mike and Pasupat, Ice and Sachdeva, Noveen and Dhillon, Inderjit and Blistein, Marcel and Ram, Ori and Zhang, Dan and Rosen, Evan and others},
  journal={arXiv preprint arXiv:2507.06261},
  year={2025}
}

@article{phi-35-vision,
  title={Phi-4 technical report},
  author={Abdin, Marah and Aneja, Jyoti and Behl, Harkirat and Bubeck, S{\'e}bastien and Eldan, Ronen and Gunasekar, Suriya and Harrison, Michael and Hewett, Russell J and Javaheripi, Mojan and Kauffmann, Piero and others},
  journal={arXiv preprint arXiv:2412.08905},
  year={2024}
}

@misc{ernie,
      title={ERNIE 4.5 Technical Report},
      author={Baidu ERNIE Team},
      year={2025},
      eprint={},
      archivePrefix={arXiv},
      primaryClass={cs.CL},
      url={}
}

@inproceedings{vllm,
  title={Efficient Memory Management for Large Language Model Serving with PagedAttention},
  author={Woosuk Kwon and Zhuohan Li and Siyuan Zhuang and Ying Sheng and Lianmin Zheng and Cody Hao Yu and Joseph E. Gonzalez and Hao Zhang and Ion Stoica},
  booktitle={Proceedings of the ACM SIGOPS 29th Symposium on Operating Systems Principles},
  year={2023}
}

@misc{lmdeploy,
    title={LMDeploy: A Toolkit for Compressing, Deploying, and Serving LLM},
    author={LMDeploy Contributors},
    howpublished = {\url{https://github.com/InternLM/lmdeploy}},
    year={2023}
}

@misc{paddle,
    title={PaddlePaddle: PArallel Distributed Deep LEarning: Machine Learning Framework from Industrial Practice},
    author={PaddlePaddle Contributors},
    howpublished = {\url{https://github.com/PaddlePaddle/Paddle}},
    year={2017}
}
\newpage



\appendix
\section{More Details about CrossVid}
\subsection{Statistical Details}

In this section, we present more statistical details of our proposed CrossVid.

\noindent \textbf{Video length for each query}
Though the number of videos involved in each query varies in CrossVid, we keep each query referring to a group of at least 2 videos, shown in Table~\ref{table_video_num}. Each query requires MLLM to reason over a total of 770 seconds of video content.

\begin{table}[!htbp]
\centering
\small
\begin{tabular}{cc}
\toprule[1pt]
\multicolumn{1}{c}{\textbf{\#Videos in the query}} & \textbf{Number of queries} \\
\hline
2                                         &     4,531              \\
3                                         &     1,665              \\
4                                         &     2,416              \\
$>$4                                  &     403              \\ \hline
Total                                     &     9,015   \\
\bottomrule[1pt]
\end{tabular}
\caption{Distribution of queries containing different numbers of videos.}
\label{table_video_num}
\end{table}

\noindent \textbf{Task statistics}
CrossVid covers 10 distinct tasks. 
The source videos for tasks are curated from six publicly available datasets: Animal Kingdom, MovieChat-1K, YouCook2, VisDrone, Charades, and Assembly101. 
We manually assign suitable videos for each type of task, and one source video can be used in different queries. 
More details, including the number of QA pairs, the number of videos for each query, and the video source, are presented in Table~\ref{tabel_task}.

\begin{table}[htbp]
\centering
\setlength{\tabcolsep}{2mm}
\small
\begin{tabular}{cccc}
\toprule[1pt]
\textbf{Task} & \textbf{\#QA pairs} & \textbf{\#Videos} & \textbf{Video sources}           \\ \hline
BU   &     848               & 3 or 4                & \makecell{Charades \&\\Animal Kingdom} \\\hline
NC   &     1,221               & 4                     & MovieChat-1K            \\\hline
CC   &     798               & 4                     & YouCook2                \\\hline
PEA  &     953               & 3                     & Assembly101             \\\hline
PI   &     251               & 2                     & MovieChat-1K            \\\hline
FSA  &     2,241               & 2                     & YouCook2                \\\hline
PSS  &     664               & 3$\sim$6              & YouCook2                \\\hline
MSR  &     595               & 2                     & VisDrone                \\\hline
MOC  &     571               & 2                     & VisDrone                \\\hline
CCQA &     873               & 2                     & YouCook2                \\
\bottomrule[1pt]
\end{tabular}
\caption{Number of QA pairs, number of videos in each QA pair, and video sources for each task.}
\label{tabel_task}
\end{table}




\subsection{Task Definition}
We give the description of the four dimensions and the detailed definition of each task in CrossVid.
The question format and example question for each task are presented in Table~\ref{table_example_questions}.

\noindent\textbf{Comparative Analysis}
This dimension evaluates the ability of MLLMs to extract task-relevant information from multiple videos and perform comparisons. It consists of the following four tasks:

\textbf{Behavioral Understanding (BU)}: A set of videos depicting either wildlife behaviors or everyday human activities is provided. For animal behavior, the model needs to recognize specific actions and understand their aims and purposes. For human activities, the model is required to accurately identify whether each video contains the queried action.

\textbf{Narrative Comprehension (NC)}: This task requires models to analyze four film clips with the same genre to contrast plot, characters, environment, and themes. 

\textbf{Culinary Comparison (CC)}: Given a group of videos showing the cooking of the same dishes, this task requires the model to compare ingredient processing, utensil usage, procedural sequence, and flavor across videos.

\textbf{Procedural Error Analysis (PEA)}: Models are provided with videos recording the assembly of the same toy car, accompanied by the descriptions of a predefined set of possible errors. Models are required to identify the errors mentioned in the question and to further trace the reasons for the mistakes.

\noindent\textbf{Temporal Understanding}
This dimension assesses the capability of models to perform temporal location and reasoning across multiple videos.
It contains the following three tasks:

\textbf{Plot Inference (PI)}: Given the beginning and ending segments of a film, the model is asked to infer the plot in the middle part. This task evaluates the ability to reason logical dependencies and causal relationships within a narrative context.

\textbf{Functional Step Alignment (FSA)}: Two cooking videos are provided, and models are asked to locate segments in one video that correspond to a specified interval in the other. This task requires aligning corresponding steps across videos and understanding semantic equivalence.

\textbf{Procedural Step Sequencing (PSS)}: A cooking video is segmented at the step level, and the clips are randomly shuffled. Models are required to reconstruct the correct temporal sequence. This task evaluates the causal reasoning and temporal inference capabilities.

\noindent\textbf{Multi-view Reasoning}
This dimension provides models with two temporally synchronized road videos, each captured from a different aerial drone. It consists of two tasks that evaluate models' cross-perspective reasoning and spatial understanding capabilities.

\textbf{Multi-view Spatial Reasoning (MSR)}: Models are queried about spatial relationships, such as relative distances and positions of objects at a specific moment, thereby evaluating multi-view spatial reasoning abilities.

\textbf{Multi-view Object Counting (MOC)}: Models are required to count objects at a specific moment or over an interval. It requires multi-perspective information integration for precise counting.

\noindent\textbf{Free-form QA}
This dimension evaluates the model’s ability to perform comparative analysis and provide comprehensive, accurate answers to open-ended questions.

\textbf{Comparative Culinary QA (CCQA)}: Two cooking videos with the same dishes are provided. Models are required to compare and identify differences in cooking procedures between the videos. This task assesses the capability to compare details.

\begin{table*}[htbp]
\centering
\setlength{\tabcolsep}{1mm}
\small
\begin{tabular}{ccc}
\toprule[1pt]
\textbf{Task}  & \textbf{Qestion Format}  & \textbf{Example Question} \\ \hline
\rowcolor{gray!10}
\multicolumn{3}{l}{Comparative Analysis}   \\
BU    & MC              &  \textit{Which cooling method in the following videos prevents water loss?}                \\
NC    & SC              &  \textit{In which video is a vehicle's role least critical to the main conflict?}               \\
CC    & SC              &  \textit{What distinguishes the final seasoning step in Video 4 compared to others?}                \\
PEA   & SC              &  \textit{Which action is incorrectly performed in exactly two videos?}                \\
\rowcolor{gray!10}
\multicolumn{3}{l}{Temporal Understanding} \\
PI    & SC              &   \textit{What is most likely to happen in the middle part?}               \\
FSA   & CG              &   \textit{Which step in Video 2 is functionally equivalent to the step shown between 57s and 68s in Video 1?}               \\
PSS   & CG              &   \textit{What is the correct order of these video segments?}               \\
\rowcolor{gray!10}
\multicolumn{3}{l}{Multi-view Reasoning}   \\
MSR   & SC              &    \textit{When $obj_1$ completely leaves view B, where is $obj_2$ located in view A's frame?}              \\
MOC   & SC              &    \textit{When $obj_1$ is parallel to the red bus, how many cars are moving in view A?}              \\
\rowcolor{gray!10}
\multicolumn{3}{l}{Free-form QA}           \\
CCQA  & OG              &    \textit{How do the two videos differ in their methods of cooking the chickpeas for chana masala?}              \\
\bottomrule[1pt]
\end{tabular}
\caption{Question formats and example questions for each task in CrossVid. The question formats include single-choice (SC), multiple-choice (MC), closed-ended generation (CG), and open-ended generation (OG).}
\label{table_example_questions}
\end{table*}

\begin{table}[htbp]
\centering
\small
\begin{tabular}{ccc}
\toprule[1pt]
\textbf{Models}             & \textbf{\#Frames} & \textbf{Imple.}  \\ 
\midrule
\rowcolor{gray!10}
\multicolumn{3}{l}{\textit{Closed-source Models}}   \\
GPT-4.1                     & $<50$     & API                           \\
GPT-4o                      & $<50$     & API                          \\
Doubao-1.5-VL-pro           & 256               & API                        \\
Gemini-2.5-Pro              & 128               & API                          \\
\midrule
\rowcolor{gray!10}
\multicolumn{3}{l}{\textit{Open-Source Models $\sim$ MoE}}  \\  
Kimi-VL-A3B-Thinking        & 256               & vLLM                    \\
ERNIE-4.5-VL-A3B            & 440               & PaddlePaddle         \\
\midrule
\rowcolor{gray!10}
\multicolumn{3}{l}{\textit{Open-source Models \textless 10B}}  \\
Qwen2.5-VL-7B               & 256               & vLLM                 \\
InternVL3-8B                & 128               & LMDeploy              \\
LongVA-7B-DPO               & 256               & HF                    \\
VideoLLaMA3-7B              & 180               & HF                   \\
Qwen2.5-Omni-7B             & 64                & vLLM                \\
Phi-3.5-vision              & 64                & vLLM                      \\
MiniCPM-O\ 2.6              & 128               & vLLM                   \\
MiMo-7B                     & 256               & vLLM                   \\
Video-R1-7B                 & 256               & vLLM                     \\
GLM-4.1V-9B-Thinking        & 256               & vLLM                     \\
\midrule
\rowcolor{gray!10}
\multicolumn{3}{l}{\textit{Open-Source Models $\sim$30B}} \\

Qwen2.5-VL-32B              & 256               & vLLM                     \\
InternVL3-38B               & 128               & LMDeploy                \\
\midrule
\rowcolor{gray!10}
\multicolumn{3}{l}{\textit{Open-Source Models $\sim$70B}}   \\
Qwen2.5-VL-72B              & 256               & vLLM                        \\
InternVL3-78B               & 128               & LMDeploy                 \\
LLaVA-Video-72B             & 128               & HF                      \\
LLaVA-OV-72B                & 24                & vLLM                       \\
\bottomrule[1pt]
\end{tabular}
\caption{Details of evaluated MLLMs, including the model name, the total number of input frames in each query (\#Frames), and the implementation method (Imple.).}
\label{table_appendix_model_details}
\end{table}

\section{More Details about Annotation Process}
\subsection{QA Pair Generation}
\noindent\textbf{Frame Caption}
To generate frame captions, we first extract frames from the source video.
For contextual coherence, we group the extracted frames temporally, with adjacent frames forming a group containing 2 to 8 frames. 
Specifically, if the metadata of the original dataset provides timestamps that segment the video (\emph{e.g.}, the time interval for each step in YouCook2), we group the frames accordingly within these intervals. 
Then, each group of frames is input to Qwen2.5-VL-72B for frame-level captioning. 
We design specific caption prompts for different types of videos to guide the model to focus on various details. 
For example, for cooking videos, it should focus more on ingredients, utensils, and actions. 
Metadata from the original dataset is also provided as input to assist in accurate understanding and captioning.
Figure \ref{fig:appendix_caption_cooking} demonstrates the captioning prompts for YouCook2.

\begin{figure*}[h]
\footnotesize
    \centering
        \begin{tcolorbox}[colback=gray!2!white, colframe=gray!10!black]
        \texttt{
Analyze several frames of one step from an instructional cooking video clip and generate a precise description.\\
Use the provided video's narrative of the current step (e.g., "making a bacon-egg sandwich") for contextual consistency.\\
\\
Core Requirements:\\
Analyze temporal information across frames and generate a concise description covering:\\
1) Actions: Primary motion of the chef (e.g., "Rotating steak with tongs for cross-hatch sear marks") + motion variations (grip angles, speed changes, pressure shifts)\\
2) Tool/Ingredient States: Active tools (e.g., "Cast iron skillet radiating visible heat waves") + ingredient transformations (color/texture changes, physical alterations)\\
3) Sensory Indicators: Steam patterns, bubbling intensity, surface crystallization\\
4) Temporal markers: Clock/time-lapse of cooking phases (i.e., timing control)\\
\\
Execution Guidelines:\\
1) Mandatory Scanning: Analyze all frames sequentially and output a description as a whole.\\
2) Visual Priority: Prioritize actual frame content over the narrative if conflicts arise.\\
3) Terminology: Use precise culinary terms over generic phrases ("cooking stuff").\\
4) Conciseness: Reduce redundant statements (e.g., environment) and use imperative sentences for conciseness. Keep the description within 140 words.\\
\\
Output format:\\
Output the generated description text within <description></description> tags.\\
\\
Provided information:\\
Recipe: \{RECIPE\}\\
Current step: \{STEP\}\\
\\
Input frames:\\
\\
        }
    \end{tcolorbox}
      \caption{Captioning prompt for YouCook2 dataset.}
      \label{fig:appendix_caption_cooking}
\end{figure*}

\noindent\textbf{QA Generation}
We employ Deepseek-R1 to automatically generate QA pairs. We provide the frame-level captions for each video. To ensure that the QA pairs are reasonable and challenging, we design task-specific prompts, guiding Deepseek-R1 to analyze the videos, generate QA, and output its rationale. Figures \ref{fig:appendix_generation_movie}, \ref{fig:appendix_generation_grounding}, and \ref{fig:appendix_generation_open} show the QA generation prompt for NC, FSA, and CCQA tasks.

\begin{figure*}[h]
\footnotesize
    \centering
        \begin{tcolorbox}[colback=gray!2!white, colframe=gray!10!black]
        \texttt{
Objective:\\
- Create QA pairs to test cross-video understanding, human-activity comparative reasoning.\\
- Input: Annotations for three video clips of their scene and captions.\\
- Output: 2 multiple-choice QA pairs. The question should focus on comparing their activities/motions/scenes.\\
\\
Requirements:\\
- Ensure questions require understanding across multiple videos, rather than focusing on a single video\\
- Create questions that test spatial, temporal, or causal reasoning\\
- Vary difficulty levels from simple observation to complex inference\\
- Avoid questions answerable from background knowledge alone\\
- Question types: similarity-finding or difference-spotting\\
- Question about unusual or unique elements across the videos is acceptable\\
\\
Question types to consider:\\
- Identify the common activity across multiple videos\\
- Spot differences in how similar actions are performed\\
- Compare locations/environments where activities occur\\
- Identify which videos contain a specific object or action\\
- Determine temporal relationships between actions\\
\\
Format requirements:\\
- The question should require analysis across all videos\\
- Include 4 options containing 1-3 correct answers (output a list of correct answers)\\
- Questions should involve comparison, contrast, or generalization\\
- When asking "which" questions, use plural nouns in the question, even if there is only one correct option.\\
e.g., "Which activities are xxx?", "Which actions are xxx?"
- When asking "Which videos xxx?", the options should be: A. Video 1 B. Video 2 C. Video 3 D. None of the above\\
\\
Output format:\\
Output a JSON format QA pairs wrapped within <QA></QA> tags.\\
\\
Input videos:\\
\{INPUT\}
        }
    \end{tcolorbox}
      \caption{QA generation prompt for the BU task.}
      \label{fig:appendix_generation_behavior}
\end{figure*}

\begin{figure*}[h]
\footnotesize
    \centering
        \begin{tcolorbox}[colback=gray!2!white, colframe=gray!10!black]
        \texttt{
Objective:\\
- Test cross-video understanding and comparative reasoning\\
- Input: 4 {CLASS} clips (plot + frame-level captions)\\
- Output: 3 QA pairs\\
\\
Requirements:\\
1. Integration Focus:\\
   - Compare character patterns, environments, plots, emotions, or themes across ALL 4 clips\\
   - Example: "How do rainy scenes affect characters' decisions in these stories?"\\
\\
2. Human-like Question:\\
   - Avoid complex vocabulary or abstract concepts\\
   - Ask questions as humanely as possible\\
   - Be imaginative in your questioning while keeping it grounded in reason\\
\\
3. Design of Options:\\
   - The phrasing of the questions and the design of options can be flexible\\
   - The options could be video numbers or text\\
   - Do not reveal too much information about each video\\
\\
4. Answer Structure:\\
   - 4 clear options (A-D) + explanation (why generate this QA, and why your answer is correct?)\\
   - type of the question: ["plot", "scene", "character", "theme", "emotion", "others"]\\
\\
5. Prohibited:\\
   - Philosophical/specialized knowledge\\
   - Trick questions\\
   - Partial film-specific questions (e.g., "In Video3 and Video4...")\\
\\
Output:\\
Output a JSON containing your QA pairs, and wrap it within <QA></QA> tags.\\
Input:\\
    \{INPUT\}
        }
    \end{tcolorbox}
      \caption{QA generation prompt for the NC task.}
      \label{fig:appendix_generation_movie}
\end{figure*}

\begin{figure*}[h]
\footnotesize
    \centering
        \begin{tcolorbox}[colback=gray!2!white, colframe=gray!10!black]
        \texttt{
Objective:\\
- Create a QA dataset to test cross-video understanding and comparative reasoning\\
- Input: 4 instructional cooking video clips (recipe + key steps frame-level captions)\\
- Output: 3 QA pairs\\
\\
Requirements:\\
1. Integration Focus:\\
   - Compare ingredient processing, tool usage, procedural variations, step order differences, timing control, or flavour styles across ALL 4 clips\\
   - Example: "How do chefs handle oil temperature control when pan-searing steak across videos?"\\
\\
2. Question:\\
   - Avoid complex vocabulary or sentence structure\\
   - Ask questions as humanely as possible\\
   - Require comparative reasoning\\
   - Be imaginative in your questioning format while keeping it grounded in reason\\
\\
3. Design of Options:\\
   - The phrasing of the questions and the design of options can be flexible\\
   - The options could be video numbers or text\\
\\
3. Answer Structure:\\
   - 4 clear options (A-D) + explanation (why generate this QA, and why your answer is correct?)\\
   - type of the question: ["ingredient", "tool", "procedure", "flavour", "timing", "others"]\\
   - make sure your answer is 100\% correct\\
\\
4. Prohibited:\\
   - Philosophical/specialized knowledge\\
   - Trick/subjective questions\\
   - Partial film-specific questions (e.g., "In Video3 and Video4...")\\
   - Reveal too much information about the videos in question or the options\\
\\
Output:\\
Output a JSON containing your QA pairs, and wrap it within <QA></QA> tags.\\
\\
Input:\\
recipe: \{RECIPE\}\\
captions:\\
\{CAPTIONS\}
        }
    \end{tcolorbox}
      \caption{QA generation prompt for the CC task.}
      \label{fig:appendix_generation_cooking}
\end{figure*}

\begin{figure*}[h]
\footnotesize
    \centering
        \begin{tcolorbox}[colback=gray!2!white, colframe=gray!10!black]
        \texttt{
Objective:\\
- Create QA pairs to test cross-video understanding, step comparative reasoning, and error identification/analysis in procedural tasks.\\
- Input: Annotations for three video clips (Video 1, 2, 3) showing assembly of the same toy car. Annotations include action segmentation (verb, object1, object2) and an error label if an error occurs.\\
- Output: Two single-choice QA pairs. The question should focus on identifying an error and its cause/type, requiring analysis across all videos.\\
\\
Error types + explanation:\\
1. Wrong order: This action is an ordering mistake.\\
2. The previous one is a mistake: This action is also an ordering mistake, but is caused by the preceding ordering mistakes in the context. \\
3. Shouldn't have happened: This action is unnecessary.\\
4. Wrong position: The two parts are not attached in their correct position.\\
\\
Reminder:\\
Steps to assemble the toy car might not be strictly fixed, but part of the action sequence has dependency constraints. \\
\\
Perform these steps to generate QA pairs:\\
1. Figure out the assembling logic and the reason for mistakes in the three videos.\\
2. Formulate the Question, consider:\\
- Comparative error identification for a specific action, e.g., "In which video is the first operation of assembling the wheels and chassis correct?"\\
- Identification of common correct/error steps, e.g., "Which step is wrong in all three videos during the first operation?"\\
- Cause of error identification, e.g., "Which step's error in the videos is caused by wrong action orders?"\\
- Cross-contextual order dependency, e.g., "What is the correct order to xxx?"\\
3. Other requirements:\\
- Do not reveal too much information about the error type you are asking about in the question.\\
- Reject partial video-specific questions (e.g., "In Video1 and Video2..."). All videos should be mentioned\\
\\
Output format:\\
You should output a JSON format QA pair and wrap it within <QA></QA> tags.\\
\\
Input format: Each row contains the action + mistake label\\
\{INPUT\}\\
        }
    \end{tcolorbox}
      \caption{QA generation prompt for the PEA task.}
      \label{fig:appendix_generation_assenmbly}
\end{figure*}

\begin{figure*}[h]
\footnotesize
    \centering
        \begin{tcolorbox}[colback=gray!2!white, colframe=gray!10!black]
        \texttt{
Objective: Create a video QA pair to test cross-video understanding and inference capability. \\
Input: A video's genre, plot summary, and frame captions with timestamps.\\
Output: Video's temporal segmentation (beginning/middle/ending) and a single-choice QA pair asking to infer the middle plot.\\
\\
Perform these steps:\\
1. **Structural Analysis**  \\
Thoroughly analyze the plot progression and captions to divide the video into three segments that represent:  \\
- Beginning (Setup/Initial Context)  \\
- Middle (Key Developments/Causal Pivot)  \\
- Ending (Resolution/Consequences)  \\
\\
2. **Temporal Segmentation**  \\
Output exact time ranges (start-end in seconds) for each segment with these requirements:  \\
- Non-overlapping intervals\\
- It is not necessary to cover the entire duration\\
- The middle segment must contain crucial causal developments that logically connect beginning and ending, i.e., the plot of the middle segment has a singular logical consistency\\
\\
3. **QA Generation**  \\
Create one single-choice question and answer pair focusing on inferring the middle segment's pivotal content based on the surrounding context.\\
Create 6 options containing one correct answer choice (original plot) and the other 5 plausible distractors. \\
Requirements:\\
- Distractors are inferences based on the context of the beginning and ending plots\\
- The answer choice is the most reasonable inference with unique logical consistency\\
- Clues in the answer choice should be found in the surrounding segments \\
- All 6 options must have a similar text length and the description granularity\\
- Text length difference less than 3 characters for each option\\
\\
Output a JSON-formatted response within <QA></QA> tags.\\
\\
Input:\\
    \{INPUT\}
        }
    \end{tcolorbox}
      \caption{QA generation prompt for the PI task.}
      \label{fig:appendix_generation_plot}
\end{figure*}

\begin{figure*}[h]
\footnotesize
    \centering
        \begin{tcolorbox}[colback=gray!2!white, colframe=gray!10!black]
        \texttt{
Objective:\\
- Create QA pairs to test cross-video temporal grounding and procedural alignment by comparing step similarity between two videos of the same recipe.\\
- Input: Two instructional cooking videos (Video A and Video B) with shared recipe steps, captions, and timestamps.\\
- Output: A question asking to identify the most relevant temporal segment in Video B corresponding to a reference segment from Video A.\\
\\
Stage 1: Cross-Video Step Analysis\\
1. Align procedural steps:\\
- Identify shared recipe steps between the two videos (e.g., "chop onions," "simmer sauce").\\
- Map physical state transitions (raw→chopped→cooked) and tool dependencies (knife→pan→oven) for each step.\\
- Note differences in step order, duration, or parallel execution (e.g., Video A seasons before frying, Video B seasons while frying).\\
2. Define reference-target pairs:\\
- Select a reference segment from video A (start/end timestamps) representing a critical step (e.g., "marinating meat").\\
- Identify functional equivalence in Video B (e.g., "seasoning meat" step, even if executed differently).\\
- Reference/target segments can be either a complete step or a substep in the videos. \\
\\
Stage 2: QA Generation\\
1. Template:\\
-  "In Video B, which temporal segment corresponds to the step in Video A's reference clip ({{ref\_seg}})? Focus on functional similarity (e.g., thermal process, ingredient state change)."\\
2. Variations:\\
- "Identify the earliest matching segment in Video B that achieves the same goal as Video A's reference clip ({{ref\_seg}})."\\
- "Which segment in Video B has the same causal role as the reference clip from Video A ({{ref\_seg}})?"\\
......\\
- Be imaginary in the questioning format.\\
3. Answer constraints:\\
- The correct answer must rely on procedural logic (e.g., "heating oil" must precede frying in both videos, even if timing differs).\\
- Exclude solutions based on low-level cues (e.g., similar camera angles, text overlays).\\
4. Output:\\
- Question (use "{{ref\_seg}}" instead of actual clip timestamps) + [start, end] timestamps of the reference clip from video A.\\
- Answer: [start, end] of the corresponding segment in video B.\\
\\
Stage 3: Validation Criteria\\
1. Unambiguous grounding: The target segment in video B must have one logically dominant match based on functional equivalence (e.g., both clips achieve "caramelization of sugar").\\
2. Reconfirmation: Adjust the reference/target segments more precisely using the provided captions with timestamps\\
3. Reject invalid cases: \\
- If steps are interleaved or parallelized differently without a clear functional match (e.g., Video A mixes dry/wet ingredients separately, Video B mixes all at once).\\
- If the reference clip's action is absent in video B.\\
\\
Example:\\
Reference Clip (Video A): [70s-100s] "Dissolving yeast in warm water"\\
Target Video (Video B): Full duration [0s-360s]\\
Question: "In video B, which time segment achieves the same functional purpose as the {{ref\_seg}} reference segment in video A?"\\
Answer: [125s-155s] ("Activating yeast with sugar and warm milk")\\
\\
Output a JSON-formatted response within <QA></QA> tags.\\
Input:\\
    \{INPUT\}\\
        }
    \end{tcolorbox}
      \caption{QA generation prompt for the FSA task.}
      \label{fig:appendix_generation_grounding}
\end{figure*}

\begin{figure*}[h]
\footnotesize
    \centering
        \begin{tcolorbox}[colback=gray!2!white, colframe=gray!10!black]
        \texttt{
Objective:\\
- Create a QA pair to test cross-video understanding \& procedural reasoning through cooking video restructuring and ordering tasks.\\
- Input: one instructional cooking video clips (recipe + key step descriptions + frame-level captions with timestamps)\\
- Output: rearranged video segments and a clip ordering question\\
- Usage: I will randomly shuffle your arranged segments and use your question to ask for the correct chronological order.\\
\\
Stage 1: Logical Video Restructuring\\
1. Analyze procedural flow:\\
- Identify temporal dependencies between steps and map physical state transitions (e.g., raw→chopped→cooked)\\
- Preserve original execution order while merging adjacent operations to form substeps (e.g., "seasoning after stir-frying")\\
2. Filter shortcuts:\\
- Remove overly leading durations (e.g., info text frame, frame with progress bar)\\
3. Number of clips:\\
- Final clips: 3-6 clips\\
\\
Stage 2: Ordering QA Generation\\
1. Create a question:\\
- Formulate question (e.g., "What is the correct chronological order to make [DISH NAME] based on essential cooking progression?")\\
2. Ensure solution validity:\\
- Ensure the order of the rearranged segments is uniquely logically correct\\
- If parallelizable steps exist, consider:\\
1) Rearrange cooking steps, e.g., merging them into composite segments\\
2) Reframe the question to temporal sequence (e.g., "Order these steps by their earliest possible starting time when optimizing preparation efficiency.")\\
3. Clip format:\\
- 2D array containing the [start, end] for each merged segment\\
\\
Stage 3: Validation Criteria\\
1. To answer the question requires an understanding of material state changes, tool/action dependencies, thermal/physical processes, etc.\\
2. Prohibit answering via shortcuts: on-screen text indicators, consistent camera angles/styles, etc.\\
3. Reject ambiguous cases if multiple valid sequences exist or steps lack clear precondition relationships.\\
\\
Output a JSON-formatted response within <QA></QA> tags.\\
\\
Input:\\
   \{INPUT\}
        }
    \end{tcolorbox}
      \caption{QA generation prompt for the PSS task.}
      \label{fig:appendix_generation_sort}
\end{figure*}

\begin{figure*}[h]
\footnotesize
    \centering
        \begin{tcolorbox}[colback=gray!2!white, colframe=gray!10!black]
        \texttt{
Objective:\\
- Create a QA dataset to test cross-video understanding and comparative reasoning\\
- Input: 2 instructional cooking video clips (Video A, Video B) with the same recipe and their key steps frame-level captions\\
- Output: 2 open-ended QA pairs with scoring criteria\\
\\
Requirements:\\
1. Integration Focus:\\
   - Compare ONE of the following aspects across both clips: ingredient processing, tool usage, procedural variations, step order differences, or flavour styles\\
   - Example: "What are the key differences in how butter is incorporated into the dish between the two videos?"\\
\\
2. Question:\\
   - Avoid complex vocabulary or sentence structure\\
   - Ask questions as humanely as possible\\
   - Require direct comparison of both videos\\
   - Focus on observable actions/decisions (no subjective interpretations)\\
   - Question format: Open-ended\\
   - Video reference: "video A" \& "video B"\\
   - Be imaginative in your questioning format while keeping it grounded in reason\\
\\
3. Answer Structure:\\
   - Single paragraph answer (3-5 sentences) summarizing comparative analysis\\
   - type of the question: ["ingredient", "tool", "procedure", "flavour", "others"]\\
   - Make sure your answer is correct\\
\\
4. Scoring Points:\\
   - 3-5 concise bullet points derived from key arguments in the answer\\
   - Each scoring point must represent an independent evidence dimension (aspect) without overlap\\
   - Scoring points cover all aspects mentioned in the answer\\
\\
5. Validation Rules:\\
   - Answers must be fully supported by video content\\
   - Philosophical/specialized knowledge \& trick/subjective questions\\
   - Comparisons must address BOTH videos equally\\
   - Partial video-specific questions (e.g., "In Video A...")\\
\\
Output:\\
Output a JSON containing your QA pairs, and wrap it within <QA</QA> tags.\\
Input:\\
recipe: 
\{RECIPE\}\\
captions:\\
\{INPUT\}\\
}
    \end{tcolorbox}
      \caption{QA generation prompt for the CCQA task.}
      \label{fig:appendix_generation_open}
\end{figure*}

\subsection{Manual Annotation}
\noindent\textbf{Multi-view Reasoning Annotation}
For both MSR and MOC tasks, we adopt a fully manual annotation approach to ensure the quality of the QA pairs. Since the objects in the videos are often small and the questions require precise spatial relationships between objects, relying solely on coarse captions cannot support fine-grained annotation.

The videos for these two tasks are sourced from the VisDrone dataset, which contains 44 pairs of synchronized drone-captured road scene videos, as well as per-frame position information (\emph{i.e.}, coordinates of the bounding boxes) for all objects in each view.

The annotation process is as follows. Firstly, for each group of videos in the VisDrone dataset, we randomly select five objects to form an object combination. For each group of videos, 100 object combinations are generated through random sampling. Next, using the per-frame bounding box information provided by the dataset, we mark these five objects in each combination with different colors on the video frames to facilitate the annotators' identification. Subsequently, annotators watch the marked videos and filter out the combinations where objects are hard to distinguish, including their colors, orientation, \emph{etc}. Finally, for the retained object combinations, annotators manually generate the questions, options, and ground truth answers. The whole process is conducted under a detailed annotation guideline that we provide to them.

The annotation guidelines include detailed restrictions and example QA pairs. The guidelines for MSR and MOC tasks are shown in Figures \ref{fig:appendix_gen_position} and \ref{fig:appendix_gen_count}, respectively.

\begin{figure*}[h]
\footnotesize
    \centering
        \begin{tcolorbox}[colback=gray!2!white, colframe=gray!10!black]
        \texttt{
1. Task Definition\\
You are required to formulate a QA pair. It should pertain to spatial relations, positions, distances, and trajectories of objects, based on paired synchronized drone videos (View A and View B). \\
The goal is to evaluate multi-view spatial reasoning ability.\\
\\
2. Question Types\\
- Relative Position/Direction\\
  - Static position: e.g., front, back, left, right, front-left, rear-right, etc. (Use the object’s moving direction as the basis; if the object is stationary and its head direction is ambiguous, do not define left/right.)\\
  - Absolute direction: e.g., north, south, east, west. If used, specify the assumption (e.g., drone moving north).\\
  - On-screen position: e.g., top-left, center, right edge of the video frame.\\
- Dynamic Position/Event Localization\\
  - Action-based descriptions: e.g., "aligned with the bicycle," "just passed the bus," "reaching under the traffic light pole."\\
- Relative Distance\\
  - Approximate quantification: "about 20 meters," "one car’s length," "three lamp post intervals."\\
- Trajectory Reasoning\\
  - Route options: e.g., go straight, turn left, turn right, stop, etc.\\
\\
3. Referencing and Annotation Rules\\
- Object Reference\\
  - Use "View+Index," e.g., \{A1\}, \{B4\}. Number/color correspondence: 1:red, 2:green, 3:blue, 4:yellow, 5:magenta.\\
  - Enclose in curly brackets. No spaces or extra characters.\\
- Special Objects/Landmarks\\
  - Must be uniquely and clearly described, e.g., "the red truck turning right," "tricycle at bottom-left corner."\\
  - Landmarks use their names ("roundabout", "gas station").\\
- Event/Time Point\\
  - Use relative descriptions, e.g., "at the beginning of the video," "when the red light turns on," "when the white car leaves the frame." Avoid absolute timestamps.\\
- View Specification\\
  - Use "View A" and "View B" consistently.\\
\\
4. Requirement for Multi-view Reasoning\\
- Questions must involve information from both views/provide reasoning that cannot be solved from a single view only.\\
- Design questions that utilize differences in occlusion between views.\\
- If a question is answerable from a single view, it is invalid for this task and should be removed.\\
\\
5. Example Templates\\
- Spatial Relation Example\\
  Q: "When \{A1\} is overtaking \{A2\} in View A, what is the approximate position of \{B3\} relative to \{A1\} in View B?"\\
  Options: (A) Front-left of \{A1\} (B) Just aligned with \{A1\} (C) Has already passed \{A1\} (D) \{A1\} not visible in View B\\
\\
- On-screen Position Example\\
  Q: "When \{A2\} in View A reaches the center of the frame, where is \{B4\} in View B?"\\
  Options: (A) Top-right (B) Bottom edge (C) Center (D) Not in View B\\
\\
- Distance Example\\
  Q: "When the white van approaches the crosswalk in View A, what is its distance to the traffic light ahead?"\\
  Options: (A) One car’s length (B) About 100 meters (C) Almost at the pole (D) Not visible in the frame\\
\\
- Trajectory Example\\
  Q: "What is the path of \{A1\}?"\\
  Options: (A) Goes straight through the intersection (B) Turns right (C) Turns left (D) Stays stationary
        }
    \end{tcolorbox}
      \caption{Guidelines for MSR manual annotation.}
      \label{fig:appendix_gen_position}
\end{figure*}

\begin{figure*}[h]
\footnotesize
    \centering
        \begin{tcolorbox}[colback=gray!2!white, colframe=gray!10!black]
        \texttt{
1. Task Definition\\
Based on paired drone views and annotated objects, formulate object counting single-choice questions involving different moments, events, or joint counting across both views.\\
The goal is to evaluate multi-view object counting ability.\\
\\
2. Question Types\\
- Single Time-point Counting\\
  E.g., "At the moment the green light turns on, how many cars are in front of the white van in its lane?"\\
- Interval Counting\\
  E.g., "During the period {A1} overtakes the pedestrian, how many white cars pass through the intersection?"\\
- Multi-view Joint Counting\\
  E.g., "How many stationary cars are visible in both View A and View B?"\\
\\
3. Counting Rules\\
- Only count objects that are clear, distinct, and unambiguously visible.\\
- For small quantities, option intervals should be close (e.g., [3, 4, 5, 6]); for large quantities, options can be spaced farther apart (e.g., [12, 14, 16, 18]).\\
\\
4. Referencing Rules\\
- Follow the same rules for object, view, and time point referencing as in MSR.\\
\\
5. Example Templates\\
- Q: "At the start of the video, how many cars are ahead of \{A1\}?"\\
  Options: (A) 3 (B) 4 (C) 5 (D) 6\\
\\
- Q: "From the beginning to the end of the video, how many vehicles does \{A3\} overtake?"\\
  Options: (A) 2 (B) 4 (C) 6 (D) 8\\
\\
- Q: "How many parked cars are visible in both View A and View B?"\\
  Options: (A) 3 (B) 4 (C) 5 (D) 6\\
        }
    \end{tcolorbox}
      \caption{Guidelines for MOC  manual annotation.}
      \label{fig:appendix_gen_count}
\end{figure*}

\subsection{Manual Review}
For the QA pairs of the remaining 8 tasks, which are generated automatically, rigorous manual review, including data filtration, refinement, and quality control, is performed. For each task, we design specific guidelines for human annotators, which are described as follows:

\textbf{Behavioral Understanding}: During filtration, questions are retained only if they are clear, objective, and require analysis across multiple videos. Those that feature strong answer cues, insufficient reasoning, or can be solved using information from a single video are discarded. In the refinement stage, annotators further revise questions to eliminate ambiguity, ensuring that each one demands meaningful behavioral comparison and multi-step reasoning based on the collective video set.

\textbf{Narrative Comprehension}: During filtration, only clear and objective questions that require reasoning across clips are retained, while subjective and ambiguous questions are removed. In the refinement, annotators ensure each single-choice question has one clear, well-supported correct answer, and all false options are plausible, so that effective narrative understanding relies on cross-clip analysis.

\textbf{Culinary Comparison}: During filtration, only clear, factual, and objective questions that require watching the videos are retained. In the refinement, annotators ensure each question has one clear, unique answer based solely on observable video content.

\textbf{Procedural Error Analysis}: During filtration, only questions that focus on error identification and reasoning, and that require careful viewing of all relevant videos, are retained. In contrast, questions answerable solely through text or focused on overly minor details are removed. In refinement, annotators ensure questions are clearly worded with precise references to the relevant assembly steps.

\textbf{Plot Inference}: During filtration, questions that cannot be answered by textual cues alone and require logical reasoning based on both the beginning and ending video segments are retained. In refinement, annotators ensure that all options share a similar text length and description granularity.

\textbf{Functional Step Alignment}: During filtration, only questions in which the reference and target intervals exhibit strong functional, contextual, or causal alignment are retained. In refinement, annotators ensure that each aligned interval captures a distinct and coherent step. We accept higher-level conceptual equivalence, such as both being seasoning steps despite using different specific ingredients.

\textbf{Procedural Step Sequencing}: During filtration, only QA pairs whose video segments have a unique correct order are retained. Those containing parallelizable steps are discarded.
In refinement, annotators realign the segments to avoid explicit guiding cues such as on-screen progress bars, step labels, and consistent camera angles.

\textbf{Comparative Culinary QA}: During filtration, only objective and factual questions that directly require cross-video comparison are retained. 
In refinement, annotators ensure that answers are accurate, fully address the question, and cover all relevant scoring points derived from the video content, with no omissions or unsupported elements.

\begin{figure*}[h]
    \centering
    \begin{subfigure}[b]{0.4\linewidth}
        \includegraphics[height=4cm]{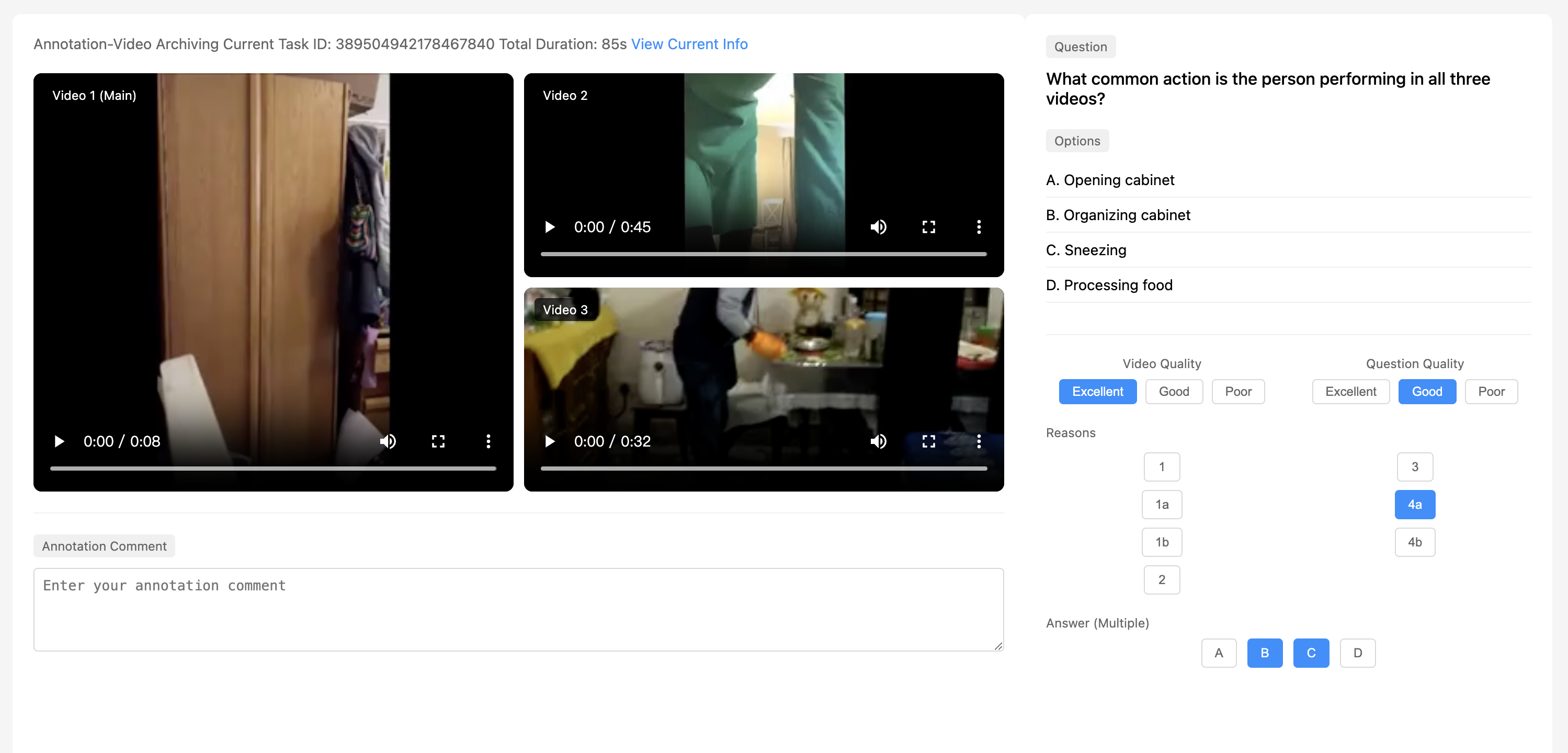}
        \caption{Annotation interface.}
        \label{fig:interface}
    \end{subfigure}
    \hfill
    \begin{subfigure}[b]{0.4\linewidth}
        \includegraphics[height=4cm]{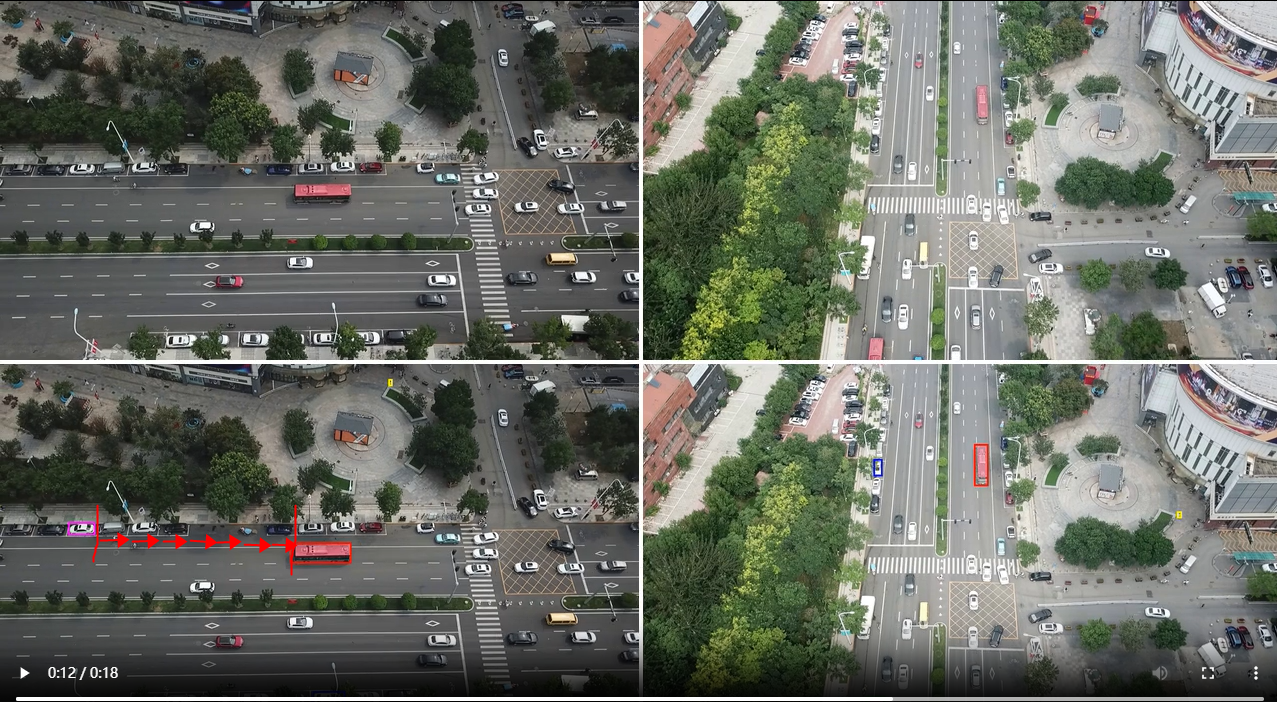}
        \caption{Marker tools for precise annotation.}
        \label{fig:marker}
    \end{subfigure}
    \caption{The interface developed for manual annotation and review to construct CrossVid.}
    \label{fig:benchmark_anno_tool}
\end{figure*}

To facilitate the manual review, we develop an interface shown in Figure~\ref{fig:interface}. Annotators can simultaneously watch the queried videos in a QA pair on the interface. The interface automatically allocates the QA pairs to each annotator and records whether the pair is discarded and the reasons. We also integrate the marker tools in the interface, facilitating precise annotation for MSR and MOC tasks, which is shown in Figure~\ref{fig:marker}.

\section{Experimental Details}

In this section, we describe more experimental details, including the evaluation metric, MLLM selection, and the hyperparameters for the experiments. 
\subsection{Evaluation Metrics}
We use accuracy to reflect the performance of MLLMs. 

For single-choice questions with only one correct option, the accuracy is calculated as the percentage of correctly answered questions.

For multiple-choice questions with one to three correct options, the model's response is correct only if it completely matches the ground truth options. For example, if the ground truth is ``AB", then the answers ``A" and ``ABC" are incorrect. Similarly, the accuracy is calculated as the percentage of correctly answered questions.

For the PSS task, MLLMs are required to output the correct order of the video segments. The response from MLLMs is correct only if the number of video segments matches the ground truth at the corresponding position.

For the FSA task, MLLMs are required to output a time interval with beginning and ending timestamps. The accuracy is calculated by the Intersection over Union (IoU), which can be expressed by:

\begin{equation}
\mathrm{IoU} = \frac{
    \max\left(0,\, \min(A_\mathrm{end},\, G_\mathrm{end}) - \max(A_\mathrm{start},\, G_\mathrm{start})\right)
}{
    \max(A_\mathrm{end},\, G_\mathrm{end}) - \min(A_\mathrm{start},\, G_\mathrm{start})
}
\label{iou}
\end{equation}
where $\left[A_\mathrm{start},A_\mathrm{end}\right]$ denotes the model's output and $\left[G_\mathrm{start},G_\mathrm{end}\right]$ denotes the ground truth time interval.

For the open-ended CCQA task, we employ GPT-4.1 to score the answer from the MLLM. We provide GPT-4.1 the MLLM's answer, the question, the scoring points, and the standard answer, and it is required to assess the response in two stages. First, for each scoring point provided in the QA pair, GPT-4.1 is required to check whether the model's answer covers the point; if so, it receives one point, otherwise zero. Then, for those scoring points regarded as covered, GPT-4.1 further evaluates whether its details exactly match those in the standard answer. If so, an additional point is added. Finally, the model's overall score is calculated as the sum of coverage and accuracy points, divided by twice the number of scoring points. The assessment prompt for GPT-4.1 is shown in Figure~\ref{fig:appendix_assess_open}.

\begin{figure*}[h]
\footnotesize
    \centering
        \begin{tcolorbox}[colback=gray!2!white, colframe=gray!10!black]
        \texttt{
You are asked to score the output of a model, given the following information:\\
- Question: {QUESTION}\\
- Standard Answer: \{ANSWER\}\\
- Scoring Points: \{POINTS\}\\
- Model's Output: \{OUTPUT\}\\
\\
Please perform the following two-part scoring:\\
Part 1: Coverage of Scoring Points\\
- For each scoring point, determine whether it is covered by the model's output.\\
- Mark as covered (true) only if the scoring point is addressed explicitly and clearly.\\
- If the mention is vague, partial, or ambiguous, consider it not covered.\\
\\
Part 2: Accuracy of Details\\
- For each covered scoring point, compare the details in the Model's Output to the Standard Answer.\\
- Mark as correct (true) only if the details are fully accurate and consistent with the Standard Answer, without any error, omission, or ambiguity.\\
- If the answer is partially correct, too broad/narrow, or not strictly consistent, mark it as not correct (false).\\
- For scoring points not covered, mark as incorrect.\\
\\
Format your answer in a JSON format as follows:\\
\{\\
\hspace*{4ex} "coverage": [true, false, true, ...],\\
\hspace*{4ex} "correctness": [true, false, false, ...]\\
\}\\
The length of "coverage" and "correctness" lists should match the number of scoring points.\\
Your answer:\\
        }
    \end{tcolorbox}
      \caption{Assessment prompt for CCQA.}
      \label{fig:appendix_assess_open}
\end{figure*}

\subsection{Implementation Details}
\noindent\textbf{Details of Experiment Settings}

For closed-source MLLMs, experiments are conducted using their official APIs. For open-source MLLMs, we conduct experiments on 8 Nvidia H800 (80 GB) GPUs. All models are obtained from publicly available repositories. For inference, we utilize vLLM~\cite{vllm}, LMDeploy~\cite{lmdeploy}, or PaddlePaddle~\cite{paddle} to accelerate processing when supported; for models that are not supported, we revert to their standard HuggingFace implementations.
During the inference, we keep the ``temperature" to zero for reproducibility. We also set a large value like 8192 for ``max\_tokens" to prevent answer truncations.

For video preprocessing, we use the maximum acceptable number of frames as the total frame count for each model. For each inference, we evenly allocate the total frame count to each video. For each video, frames are sampled uniformly and resized so that the longer side is 360 pixels, maintaining the aspect ratio. The details of the evaluated MLLMs are shown in Table~\ref{table_appendix_model_details}, where ``HF" denotes the official implementation on HuggingFace.

The frames of all queried videos and the question prompt are fed to the MLLM together. The video frames are input in sequence, and before each video, a text prompt indicates the MLLM of the video number. We adopt a zero-shot strategy, and the MLLM is required to produce its answer directly. 
We use ``\textit{You are a helpful video analyzer.}" as the system prompt.

For user prompts, we demonstrate the template inference prompts for single-choice questions, multiple-choice questions, the FSA task, the PSS task, and the open-ended CCQA task in Figures \ref{fig:appendix_infer_single}, \ref{fig:appendix_infer_multi}, \ref{fig:appendix_infer_grounding}, \ref{fig:appendix_infer_sort}, and \ref{fig:appendix_infer_open}, respectively. Particularly, for both MSR and MOC tasks, we refer to objects in the frame using $obj_1$, $obj_2$ ... in the question and options. Their coordinates in their first appearing frames of the referring view are provided, respectively. We use the resized bounding boxes provided in the original dataset as the coordinates. The inference prompt for MSR and MOC tasks is illustrated in Figure~\ref{fig:appendix_infer_uav}.

\begin{figure*}[h]
\footnotesize
    \centering
        \begin{tcolorbox}[colback=gray!2!white, colframe=gray!10!black]
        \texttt{
Provide you with four videos and a single-choice question with only one correct option.\\
Watch the videos carefully, and think about the question based on the information from these videos.\\
Select one answer choice, and only output the capital letter of your choice.\\
\\
Question:\\
\{QUESTION\}\\
\\
Options:\\
\{OPTIONS\}\\
\\
Input frames:\\
Video1: \{FRAMES1\}\\
Video2: \{FRAMES2\}\\
...\\
\\
Your answer:\\
        }
    \end{tcolorbox}
      \caption{Inference prompt for single-choice questions.}
      \label{fig:appendix_infer_single}
\end{figure*}

\begin{figure*}[h]
\footnotesize
    \centering
        \begin{tcolorbox}[colback=gray!2!white, colframe=gray!10!black]
        \texttt{
Provide you with four videos and a multiple-choice question with 1-3 correct answer choices.\\
Watch the videos carefully, and think about the question based on the information from the four videos.\\
Only output the capital letters of ALL your choices, e.g., "BCD".\\
\\
Question:\\
\{QUESTION\}\\
\\
Options:\\
\{OPTIONS\}\\
\\
Input frames:\\
Video1: \{FRAMES1\}\\
Video2: \{FRAMES2\}\\
...\\
\\
Your answer:\\
        }
    \end{tcolorbox}
      \caption{Inference prompt for multiple-choice questions.}
      \label{fig:appendix_infer_multi}
\end{figure*}

\begin{figure*}[h]
\footnotesize
    \centering
        \begin{tcolorbox}[colback=gray!2!white, colframe=gray!10!black]
        \texttt{
Provide you with two cooking videos, which step in Video 2 is functionally equivalent to the step shown between {BEGIN}s and {END}s in Video 1?\\
Timestamps of frames sampled from Video 1 are: \{TIME1\}.\\
Timestamps of frames sampled from Video 2 are: \{TIME2\}.\\
Watch the two videos carefully, and think about the question based on the information in the two videos.\\
Only output a time interval in seconds and separate the beginning and ending times with a comma, e.g., "15,23".\\
\\
Input frames:\\
Video1: \{FRAMES1\}\\
Video2: \{FRAMES2\}\\
\\
Your answer:\\
        }
    \end{tcolorbox}
      \caption{Inference prompt for the FSA task.}
      \label{fig:appendix_infer_grounding}
\end{figure*}

\begin{figure*}[h]
\footnotesize
    \centering
        \begin{tcolorbox}[colback=gray!2!white, colframe=gray!10!black]
        \texttt{
Provide you with \{N\} shuffled segments of a cooking video, what's the correct order of these segments?\\
Watch the segments carefully, and think about the question based on the relationship between these segments.\\
Only output the correct segment number sequence separated by "->", e.g., "2->3->1->4".\\
\\
Input frames:\\
Video1: \{FRAMES1\}\\
Video2: \{FRAMES2\}\\
\\
Your answer:\\
        }
    \end{tcolorbox}
      \caption{Inference prompt for the PSS task.}
      \label{fig:appendix_infer_sort}
\end{figure*}

\begin{figure*}[h]
\footnotesize
    \centering
        \begin{tcolorbox}[colback=gray!2!white, colframe=gray!10!black]
        \texttt{
Provide you with two cooking videos (Video A + Video B) and an open-ended question.\\
Watch the videos carefully, and think about the question based on the information from both videos.\\
\\
Question:\\
\{QUESTION\}\\
\\
Input frames:\\
Video1: \{FRAMES1\}\\
\\
Your answer:\\
        }
    \end{tcolorbox}
      \caption{Inference prompt for the open-ended CCQA task.}
      \label{fig:appendix_infer_open}
\end{figure*}

\begin{figure*}[h]
\footnotesize
    \centering
        \begin{tcolorbox}[colback=gray!2!white, colframe=gray!10!black]
        \texttt{
Provide you with two synchronized UAV road recording videos, objects' positional information, and a multiple-choice question.\\
The positional information contains the bounding box coordinates ([xtl, ytl, xbr, ybr]) of the objects positioned in the first appearing frame.\\
Watch the videos first, then track the objects in both views and think about the question based on the information.\\
Select one answer choice, and only output the capital letter of your choice.\\
\\
Question:\\
\{QUESTION\}\\
\\
Options:\\
\{OPTIONS\}\\
\\
Object information:\\
$obj_1$ first appears in the 2nd frame of view A with bbox \{BBOX1\}\\
...\\
\\
Input frames:\\
View A: \{FRAMES1\}\\
View B: \{FRAMES2\}\\
\\
Your Answer:\\
        }
    \end{tcolorbox}
      \caption{Inference prompt for MSR and MOC task.}
      \label{fig:appendix_infer_uav}
\end{figure*}

\noindent\textbf{Details of CoT}
To evaluate the effectiveness of the Chain-of-Thought (CoT) prompting, we revise the original prompts and explicitly instruct the MLLMs to generate their answers in three stages: 1) understand the question; 2) analyze the frames for each video; 3) provide the answer based on thorough analysis and double-check. We provide the CoT prompt in Figure~\ref{fig:appendix_infer_cot}.

\begin{figure*}[h]
\footnotesize
    \centering
        \begin{tcolorbox}[colback=gray!2!white, colframe=gray!10!black]
        \texttt{
Provide you with four videos and a single-choice question with only one correct option.\\
Watch the videos carefully, and think about the question based on the information from these videos.\\
\\
Follow these thinking steps to answer:\\
- Analyze the question and describe the key element in the question.\\
- Carefully observe the frames from the provided videos and briefly describe the key information.\\
- Aggregate the information and analyze each option. Explain your reasoning.\\
- Based on your analysis, select the best answer.\\
\\
You should first output the above thinking steps within <think></think> tags.\\
Then, output the capital letter of your answer choice within <answer></answer> tags.\\
\\
Question:\\
\{QUESTION\}\\
\\
Options:\\
\{OPTIONS\}\\
\\
Input frames:\\
Video1: \{FRAMES1\}\\
Video2: \{FRAMES2\}\\
...\\
\\
Your answer:\\
        }
    \end{tcolorbox}
      \caption{CoT prompt for single-choice questions.}
      \label{fig:appendix_infer_cot}
\end{figure*}

\section{More Details about Error Analysis}
\begin{figure*}[h]
\centering
\includegraphics[width=0.98\textwidth]{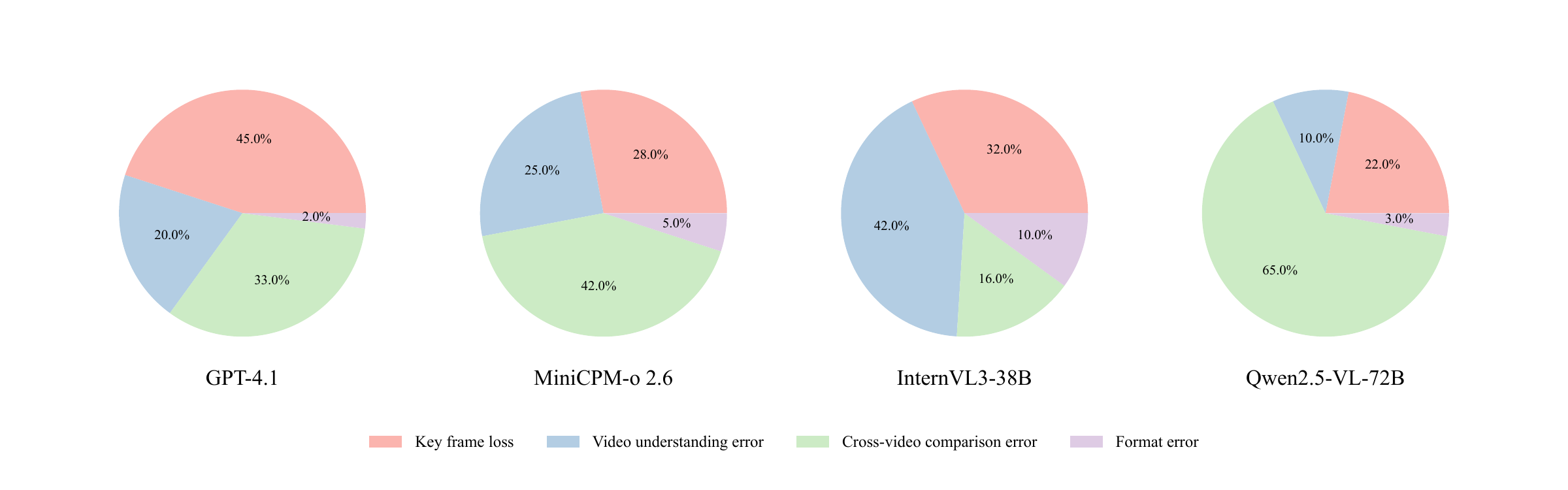}
\caption{Percentage of each error type for each MLLM.}
\label{fig_error_percentage}
\end{figure*}

We manually analyze the reasoning steps of the four model (GPT-4.1, MiniCPM-o\ 2.6, InternVL3-38B, and Qwen2.5-VL-72B) based on their responses under CoT prompting. The percentage of the four error types (key frame loss, video understanding error, cross-video comparison error, and format error) for each MLLM is shown in Figure~\ref{fig_error_percentage}. It can be observed that MLLMs with more input frames have fewer key frame errors. Most MLLMs are able to understand the single video accurately; however, they still struggle with cross-video comparison when required to process multiple videos simultaneously. 

\begin{figure*}[h]
\centering
\includegraphics[width=0.95\textwidth]{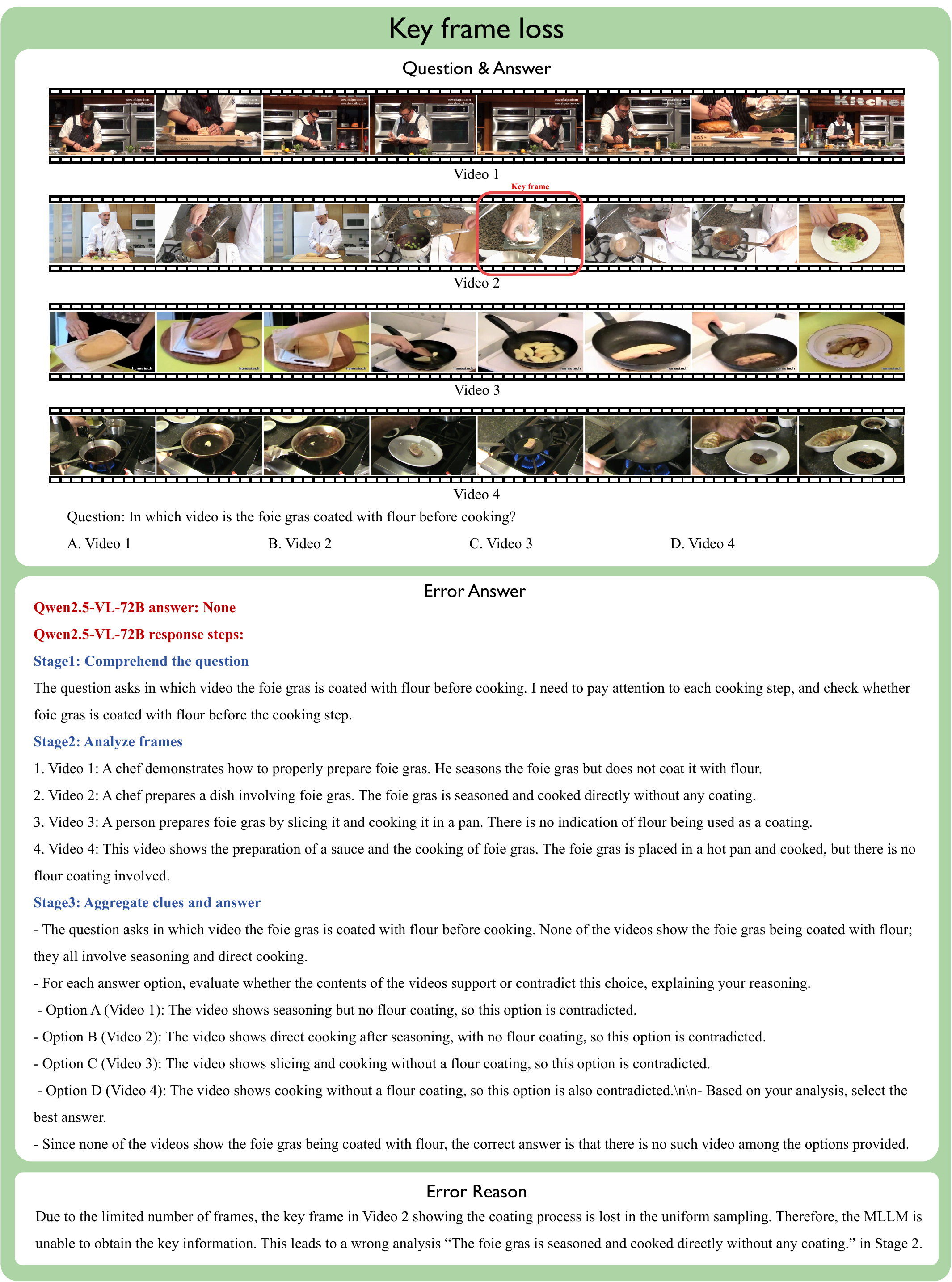}
\caption{Example of the key frame loss.}
\label{fig_key_frame_error}
\end{figure*}

For further analysis, we present visualized examples of the errors.
Figure~\ref{fig_key_frame_error} shows an example of the key frame loss. The question requires the MLLM to judge whether the foie gras is coated with flour before the cooking step for each video. It can be clearly observed that in video 2, the foie gras is coated with flour before cooking; however, the frames of coating might be missing when uniformly sampling. Hence, Qwen2.5-VL-72B is unknown in this detail and produces the wrong answer.

\begin{figure*}[h]
\centering
\includegraphics[width=0.95\textwidth]{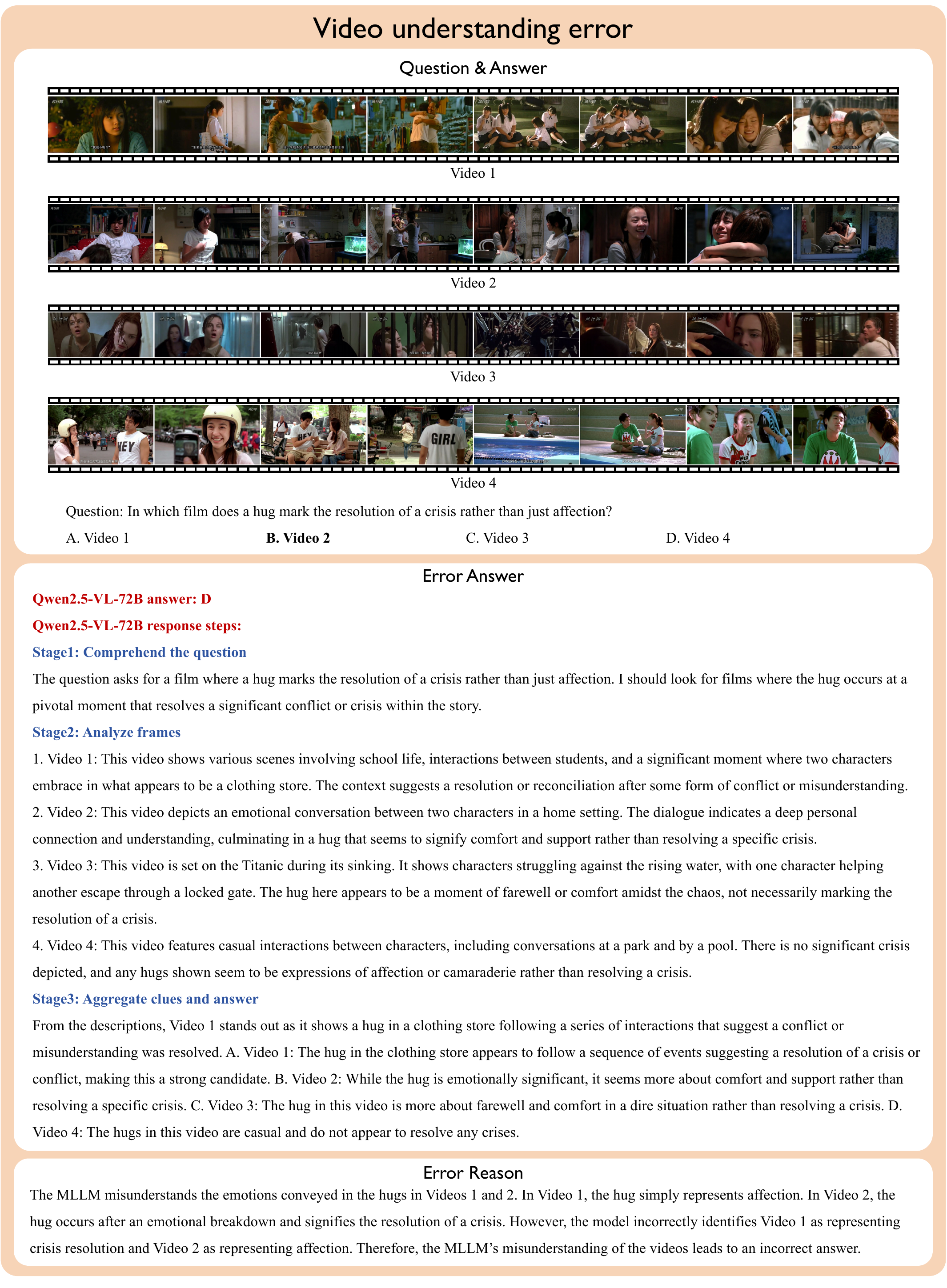}
\caption{Example of the video understanding error.}
\label{fig_understanding_error}
\end{figure*}

Figure~\ref{fig_understanding_error} shows an example of the video understanding error. The question requires the MLLM to distinguish the contextual meaning of the hugging in each video. Qwen2.5-VL-72B successfully captures the frames containing the hugging for each video, while it fails to correctly understand video 2 and thus produces the wrong answer.

\begin{figure*}[h]
\centering
\includegraphics[width=0.95\textwidth]{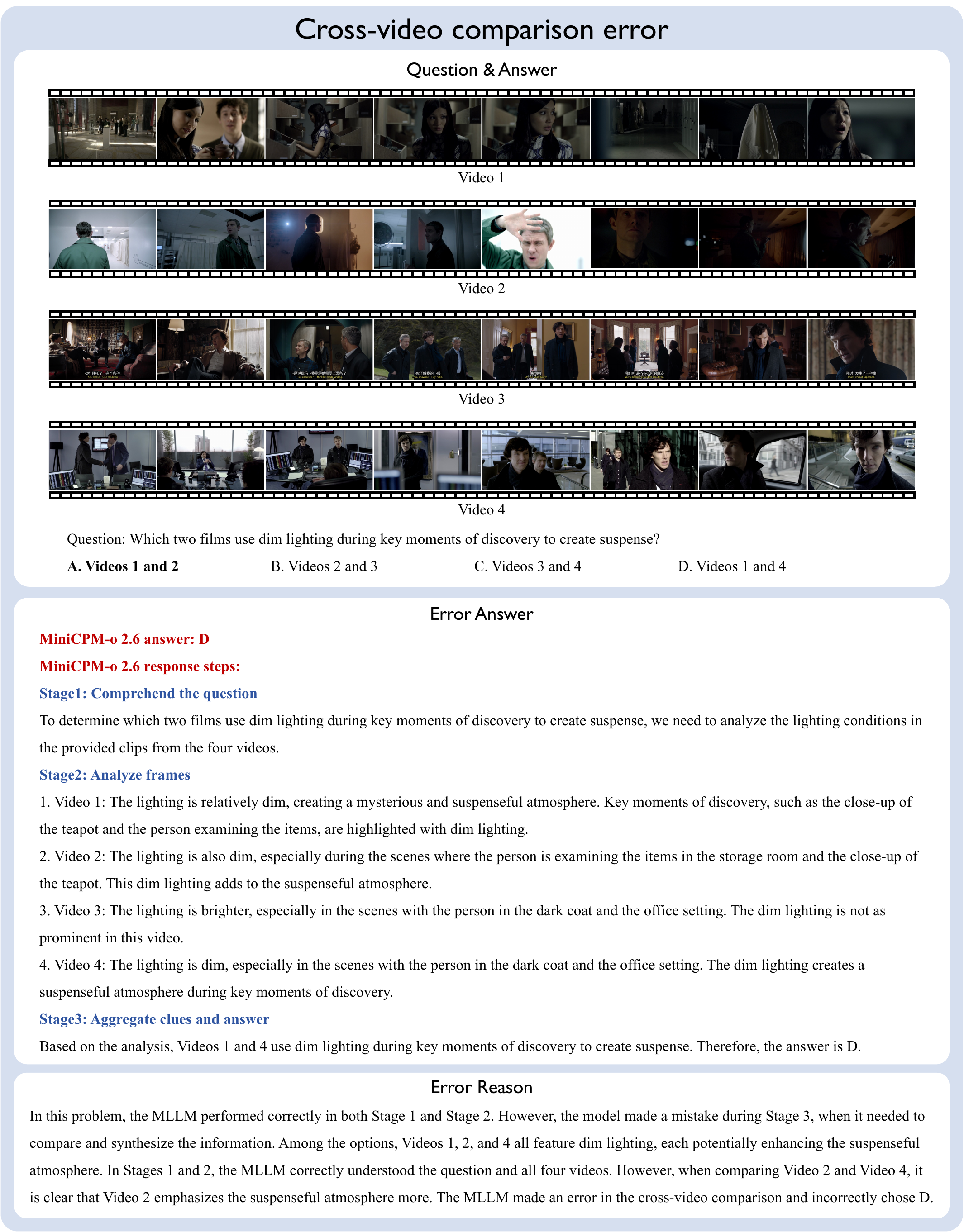}
\caption{Example of the cross-video comparison error.}
\label{fig_cross_video_error}
\end{figure*}

Figure~\ref{fig_cross_video_error} shows an example of the cross-video comparison error. The question requires the MLLM to analyze the function of dim light in creating a suspenseful atmosphere. MiniCPM-o\ 2.6 analyzes the lighting and the atmosphere in each video. However, when aggregating clues across videos, the MLLM gives its answer based on simple comparisons.  

\end{document}